\pdfoutput=1
\documentclass[pdflatex,sn-mathphys-num]{sn-jnl}
\usepackage{graphicx}%
\usepackage{multirow}%
\usepackage{amsmath,amssymb,amsfonts}%
\usepackage{amsthm}%
\usepackage{mathrsfs}%
\usepackage[title]{appendix}%
\usepackage{xcolor}%
\usepackage{textcomp}%
\usepackage{manyfoot}%
\usepackage{booktabs}%
\usepackage{algorithm}%
\usepackage{algorithmicx}%
\usepackage{algpseudocode}%
\usepackage{listings}%
\usepackage{subcaption} 

\theoremstyle{thmstyleone}%
\theoremstyle{thmstyletwo}%

\theoremstyle{thmstylethree}%

\begin{document}

\title[Article Title]{LEGO-MOF: Equivariant Latent Manipulation for Editable, Generative, and Optimizable MOF Design}

\author[1,2]{\fnm{Chaoran} \sur{Zhang}}\email{chaoranzhang@link.cuhk.edu.cn}

\author[1]{\fnm{Guangyao} \sur{Li}}\email{liguangyao@cuhk.edu.cn}

\author*[1]{\fnm{Dongxu} \sur{Ji}}\email{jidongxu@cuhk.edu.cn}

\affil[1]{\orgdiv{School of Science and Engineering (SSE)}, \orgname{The Chinese University of Hong Kong, Shenzhen (CUHKSZ)}, \orgaddress{\city{Shenzhen}, \country{China}}}

\affil[2]{\orgdiv{School of Data Science (SDS)}, \orgname{The Chinese University of Hong Kong, Shenzhen (CUHKSZ)}, \orgaddress{\city{Shenzhen}, \country{China}}}

\abstract{Metal-organic frameworks (MOFs) are highly promising for carbon capture, yet navigating their vast design space remains challenging. Recent deep generative models enable \textit{de novo} MOF design but primarily act as feed-forward structure generators. By heavily relying on predefined building block libraries and non-differentiable post-optimization, they fundamentally sever the information flow required for continuous structural editing. Here, we propose a target-driven generative framework focused on continuous structural manipulation. At its core is LinkerVAE, which maps discrete 3D chemical graphs into a continuous, SE(3)-equivariant latent space. This smooth manifold unlocks geometry-aware manipulations, including implicit chemical style transfer and zero-shot isoreticular expansion. Building upon this, we introduce a test-time optimization (TTO) strategy, utilizing an accurate surrogate model to continuously optimize the latent graphs of existing MOFs toward desired properties. This approach systematically enhances carbon capture performance, achieving a striking average relative boost of 147.5\% in pure CO$_2$ uptake while strictly preserving structural validity. Integrated with a latent diffusion model and rigid-body assembly for full MOF construction, our framework establishes a scalable, fully differentiable pathway for both the automated discovery, targeted optimization and editing of functional materials.}

\maketitle

\section{Introduction}\label{sec1}

Metal-Organic Frameworks (MOFs) are crystalline materials constructed from inorganic metal nodes and organic linkers. Their modular nature endows them with exceptional porosity, high surface areas, and tunable topologies, making them prime candidates for addressing global challenges such as carbon capture, water harvesting, and catalysis \cite{li1999design, rowsell2004metal, yusuf2022review, ma2023porous}. However, the combinatorial explosion of possible node-linker-topology combinations creates an astronomical design space \cite{yin2022computational, han2025development, comlek2023rapid, ma2026advancing}. Traditionally, the discovery of novel MOFs relies heavily on chemical intuition, heuristic rules, and high-throughput computational screening over hypothetical databases generated by algorithmic assembly tools \cite{tobacco,pormake}. Navigating this vast space requires optimizing materials beyond their initial configurations, a key paradigm in data-driven materials discovery \cite{yin2022computational,zhang2026multiobjective,wang2025accelerating}. A classic approach to such optimization relies on the principles of isoreticular chemistry, where frameworks with identical topology are systematically expanded to enlarge pore sizes and surface areas while preserving structural integrity \cite{furukawa2011isoreticular,schukraft2017isoreticular}. 

To accelerate discovery and overcome the constraints of heuristic assembly, deep generative models have recently emerged as powerful tools. For small molecules, diffusion models have achieved remarkable success in capturing both spatial topologies and chemical information \cite{edm,geodiff,difflinker}. Extending these methods to periodic crystals introduces additional complexities, pioneered by models integrating Variational Autoencoders (VAEs) with diffusion \cite{cdvae} and large-scale property-guided frameworks \cite{mattergen}. To address the uniquely large hierarchical architectures of MOFs, recent models have adopted coarse-grained or fragment-based strategies to generate skeletal arrangements \cite{mofdiff,bba}, utilized Riemannian flow matching for rigid-body assembly \cite{mofflow}, or developed fully all-atom frameworks \cite{mofasa,atomof} and agentic LLM-driven pipelines \cite{mofgen,chatmof}. 

Despite these significant advances, a critical gap remains: existing machine learning-driven approaches primarily act as feed-forward structure generators. They frequently parameterize organic linkers based on fixed building-block libraries \cite{mofdiff,xrd2mof}, which inherently breaks differentiability, discards continuous spatial and rotational gradients, and precludes the targeted editing of existing materials. To enable continuous manipulation, discrete chemical graphs must be mapped to smooth, low-dimensional latent spaces. While this has been explored using junction trees for 2D graphs \cite{cdvae}, 1D descriptors for MOF inverse design \cite{egmof}, and 3D VAEs to decouple spatial coordinates from invariant chemical features \cite{molflae}, directly optimizing structures remains a formidable challenge. Although advanced graph neural networks demonstrate state-of-the-art accuracy as property predictors \cite{cgcnn,schnet,alignn}, effectively leveraging them as differentiable surrogate models for inverse design requires a fully continuous, geometry-aware structural representation that existing MOF generators lack.

To overcome these limitations, we introduce a highly modular framework that fundamentally shifts the focus from one-shot generation to the continuous editing, targeted optimization, and zero-shot discovery of MOFs. We leverage the metal-oxo algorithm from MOFid \cite{mofid} to logically decompose complex MOF structures into constituent linkers and metal nodes. Central to our framework is \textbf{LinkerVAE}, a generative architecture inspired by MolFLAE \cite{molflae}. LinkerVAE utilizes an SE(3)-equivariant encoder to compress variable-length linker graphs into a fixed-length continuous latent representation $Z = (Z_x, Z_h)$. This explicitly disentangles 3D spatial coordinates ($Z_x$) from invariant chemical and topological features ($Z_h$), establishing a fully differentiable manifold that overcomes the rigid constraints of discrete atomic vocabularies.

This disentangled latent manifold serves as the foundation for powerful geometry-aware editing mechanisms. First, we enable \textit{Chemical Editing} via linker style transfer, manipulating invariant chemical features ($Z_h$) to interpolate functional groups across different linkers. During decoding, we can seamlessly perturb atom counts while rigorously preserving the structural validity and connection topologies of anchor atoms. Second, we achieve \textit{Physical Geometry Manipulation} via zero-shot isoreticular expansion. By preserving chemical features ($Z_h$) and selectively stretching the geometric coordinates ($Z_x$) along the principal growth axis, we systematically expand linker backbones. Coupled with gradient-based connection optimization, this allows for the precise enlargement of MOF pore sizes without requiring novel training data.

Building directly upon this continuous representation, we propose a Latent Test-Time Optimization (TTO) strategy. We integrate a periodic SchNet-based surrogate predictor operating on the coarse-grained latent crystal graph, achieving exceptional accuracy in predicting carbon capture properties. By freezing the metal nodes and applying gradient-based updates directly to the continuous latent variables ($Z_h, Z_x$) and target atom counts guided by this surrogate, we systematically evolve the structure of existing MOFs. Our experiments demonstrate that Latent TTO is remarkably robust, achieving substantial performance improvements across the chemical space, even when optimizing the highest-performing MOF candidates.

Finally, to ensure our framework remains capable of exploring completely uncharted domains, we train an equivariant Linker Latent Diffusion Model (LLDM) over the LinkerVAE latent space, adopting an equivariant graph neural network backbone. This completes a comprehensive pipeline where newly generated, highly valid linkers can be seamlessly integrated into full 3D periodic crystals using standard assembly tools \cite{tobacco, pormake} or advanced flow-matching assembly pipelines \cite{mofflow}, providing a scalable and fully differentiable pathway for automated materials discovery and targeted engineering.

\begin{figure}[htbp]
\centering
\includegraphics[width=1.0\textwidth]{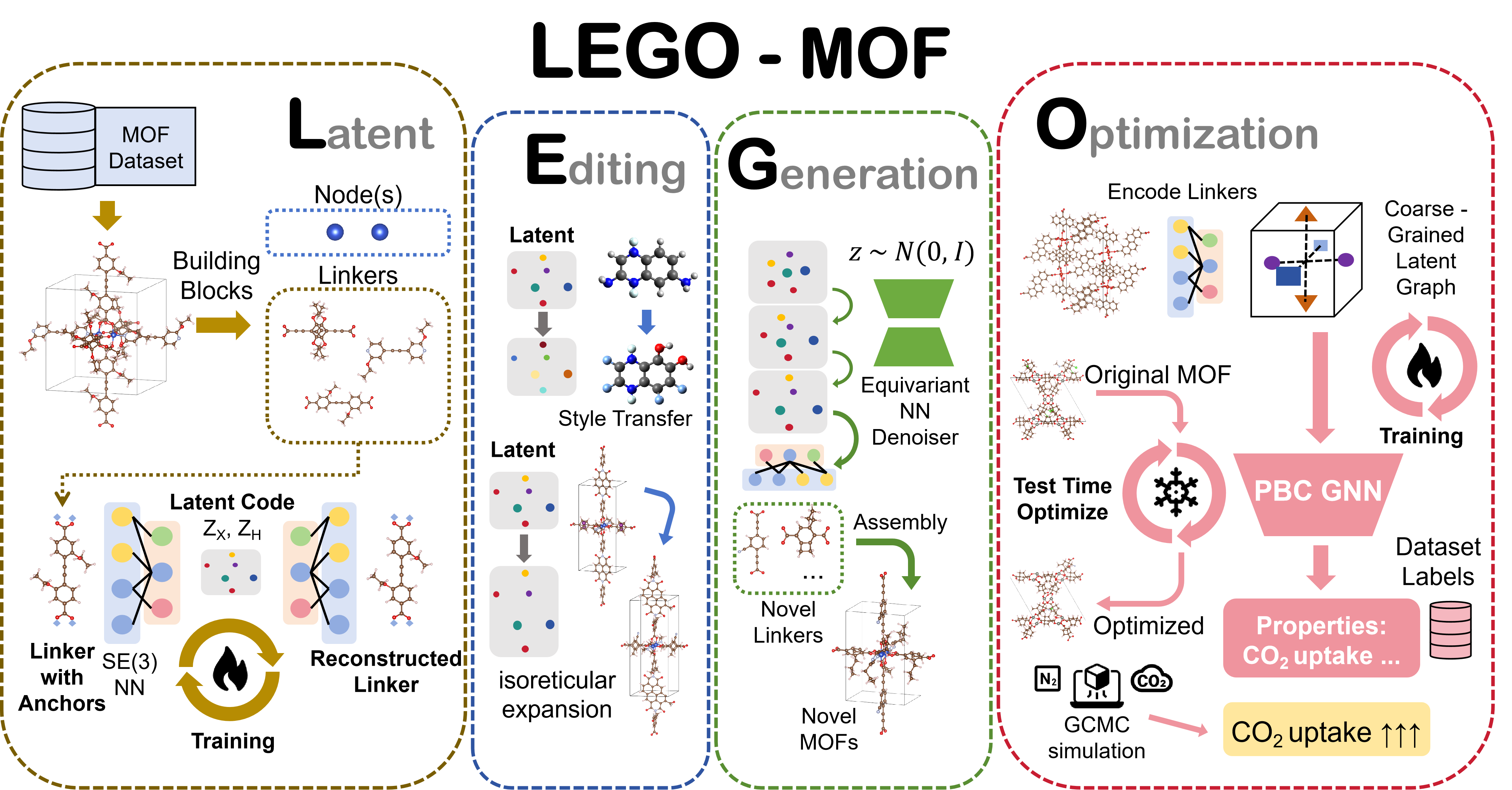}
\caption{\textbf{Overview of the LEGO-MOF framework.} The pipeline is unified by four highly modular stages: \textbf{(L)atent Representation:} Complex MOFs from databases are deconstructed into metal nodes and organic linkers. An SE(3)-equivariant autoencoder compresses variable-length linkers (including virtual anchor atoms) into a fixed-dimensional, disentangled continuous latent space ($Z_x, Z_h$). \textbf{(E)diting:} This continuous manifold enables zero-shot explicit and implicit structural manipulations, such as functional group style transfer and geometric isoreticular expansion, via targeted latent perturbations. \textbf{(G)eneration:} To explore uncharted chemical space, an equivariant latent diffusion model unconditionally samples from a Gaussian prior ($z \sim \mathcal{N}(0, I)$) to generate novel linker molecules, which are then integrated via rigid-body assembly into \textit{de novo} MOFs. \textbf{(O)ptimization:} A periodic graph neural network (PBC GNN) trained on coarse-grained latent graphs serves as a differentiable surrogate. Test-Time Optimization (TTO) leverages this surrogate to perform direct gradient ascent on the latent codes of existing MOFs, systematically maximizing target properties (e.g., CO$_2$ uptake) for downstream validation.}
\label{fig:lego_mof_overview}
\end{figure}

The overall architecture and synergistic workflows of the proposed framework—encompassing representation, editing, generation, and targeted optimization—are schematically illustrated in Figure \ref{fig:lego_mof_overview}.

\section{Results}\label{sec2}

\subsection{Continuous and Disentangled Latent Representation of Organic Linkers}

A fundamental prerequisite for manipulating MOFs is the ability to encode discrete, variable-length 3D chemical graphs into a smooth, fixed-dimensional latent space \cite{mofflow, xrd2mof}. Our LinkerVAE achieves this by explicitly disentangling the geometric coordinates ($Z_x$) from the invariant chemical and topological features ($Z_h$) while strictly preserving SE(3) equivariance. Crucially, to ensure that the generated organic linkers can be seamlessly integrated into periodic MOF structures via our rigid-body assembly pipeline, the model must meticulously recognize and preserve the structural ``anchor'' atoms (represented virtually by Helium) that serve as connection points to the metal nodes.

To evaluate the representational capacity of LinkerVAE, we first tested its ability to reconstruct unseen organic linkers from the test set (Supplementary Table 1). The model demonstrated exceptional fidelity in preserving both spatial geometry and chemical identity. Specifically, the reconstructed linkers achieved a low Root Mean Square Deviation (RMSD) of 0.566 $\pm$ 0.911 \AA~ without any external alignment (unaligned RMSD), indicating that the absolute spatial coordinates are effectively memorized by the equivariant latent variables. In terms of chemical identity, the model accurately predicted atom types with an accuracy of 94.84\% (excluding anchor atoms).

More importantly for MOF construction, LinkerVAE exhibited near-perfect retention of the anchor atoms. Across the test set, the difference in the count of anchor atoms between the ground truth and the reconstructed linkers was consistently zero, with a minor Chamfer distance spatial deviation of only 0.307 \AA. This confirms that the model robustly preserves the topological attachment points necessary for valid downstream lattice assembly.

\subsection{Robust Geometry-Aware Atom Count Manipulation}

Traditional nearest-neighbor discrete representations of building blocks preclude the generation of fundamentally novel linkers. In contrast, the fixed-dimensional nature of the LinkerVAE latent space naturally decouples the chemical encoding from the explicit number of atoms. By maintaining the encoded latent representation $(Z_x, Z_h)$ constant and perturbing the target atom count ($N_{atoms}$) during the Continuous Time Bayesian Flow Network (BFN) decoding phase, we can systematically induce structural expansions or contractions within a single molecule.

We conducted an atomic change experiment on 1000 candidate linkers (filtered for original lengths $\ge$ 8 atoms) by artificially shifting the decoded atom count by a delta ($\Delta$) ranging from -3 to +3, as demonstrated in Fig.~\ref{fig:atomic_change}. 

The result in Supplementary Table 2 shows that the model remarkably maintained a 100\% chemical validity across all perturbations. When $\Delta=0$, the system successfully reconstructed the original molecules with a low baseline RMSD of 0.279 \AA. Crucially, despite the addition or removal of heavy atoms, the geometric positioning of the anchor atoms remained highly constrained. The difference in anchor atom counts consistently averaged near zero (ranging from 0.014 to 0.052), and the spatial deviation of the anchors remained stable between 0.37 \AA~ and 0.47 \AA. This demonstrates that LinkerVAE's geometric latent variable ($Z_x$) exerts strong regularization over the overall spatial scaffold.

\begin{figure}[htbp]
\centering
\includegraphics[width=0.95\textwidth]{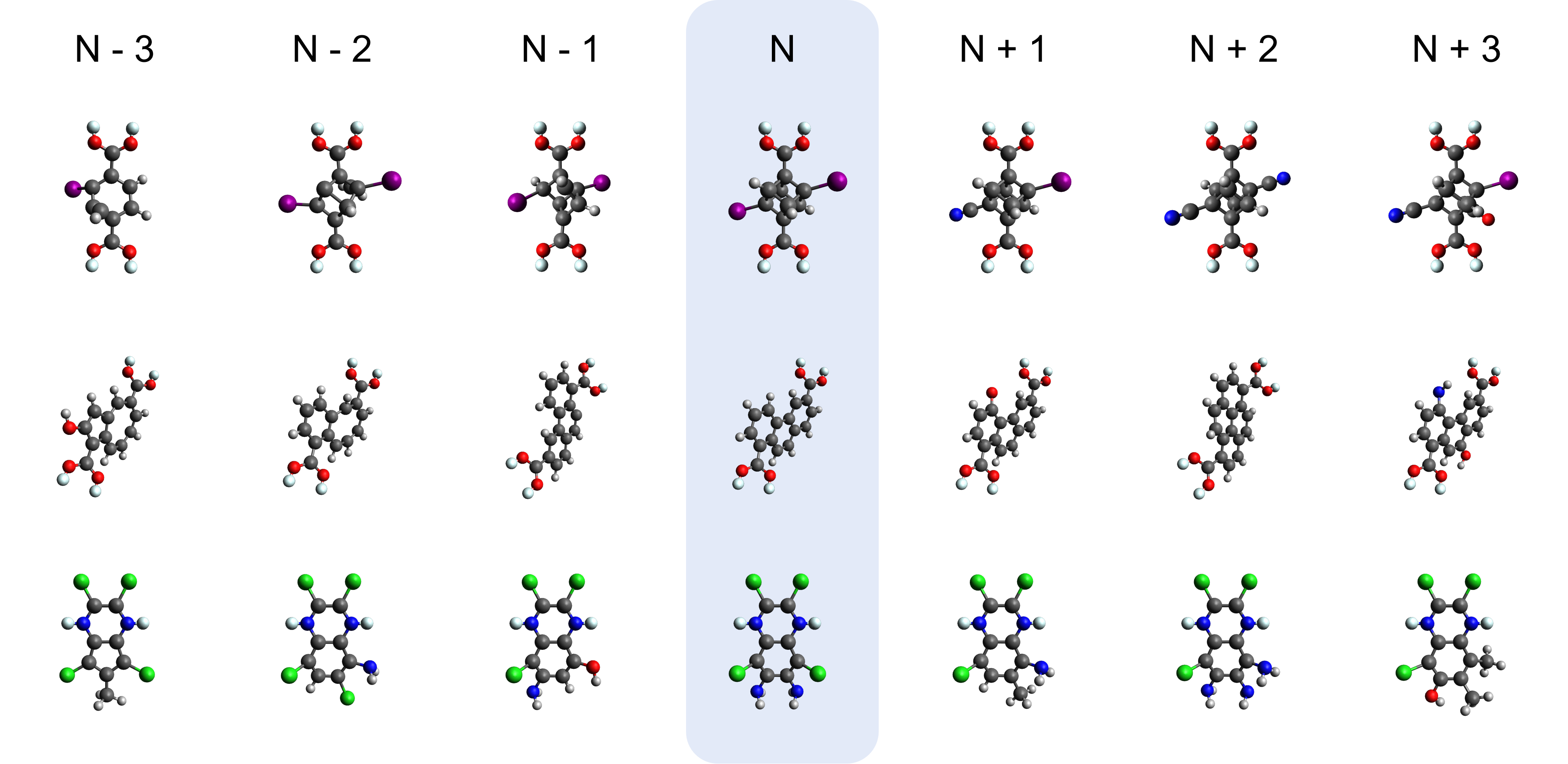}
\caption{\textbf{Continuous atomic manipulation via LinkerVAE.} Structural evolution of a representative organic linker under targeted atom count perturbations ($\Delta \in [-3, +3]$). The three rows represent three different linkers, and the light-blue atoms denote anchor sites (He).}
\label{fig:atomic_change}
\end{figure}

\subsection{Latent Style Transfer and Chemical Editing}

To further validate the successful disentanglement of physical geometry and chemical properties in our latent space, we performed a structural ``style transfer'' experiment. This paradigm is conceptually analogous to scaffold hopping in molecular design, where the underlying functional properties of a molecule are preserved while its structural motifs or functional groups are systematically altered, enabling efficient exploration of novel chemical space and lead optimization \cite{wang2025scaffold,zheng2021scaffold,hu2023scaffold}.

We randomly paired valid linkers possessing distinct chemical compositions but similar anchor distances ($<2.2$ \AA~ difference). For a given pair (P1 and P2), we linearly interpolated their chemical latent features, defining a mixed chemical style $Z_{h, mix} = 0.5 Z_{h,1} + 0.5 Z_{h,2}$. We then decoded two distinct sets of generated linkers (Gen1 and Gen2) by combining $Z_{h, mix}$ with the respective unperturbed geometric latent coordinates ($Z_{x,1}$ and $Z_{x,2}$). We tested 1000 pairs of different linkers and calculated the similarity, validity and difference in anchor atoms.

\begin{figure}[htbp]
\centering
\includegraphics[width=0.75\textwidth]{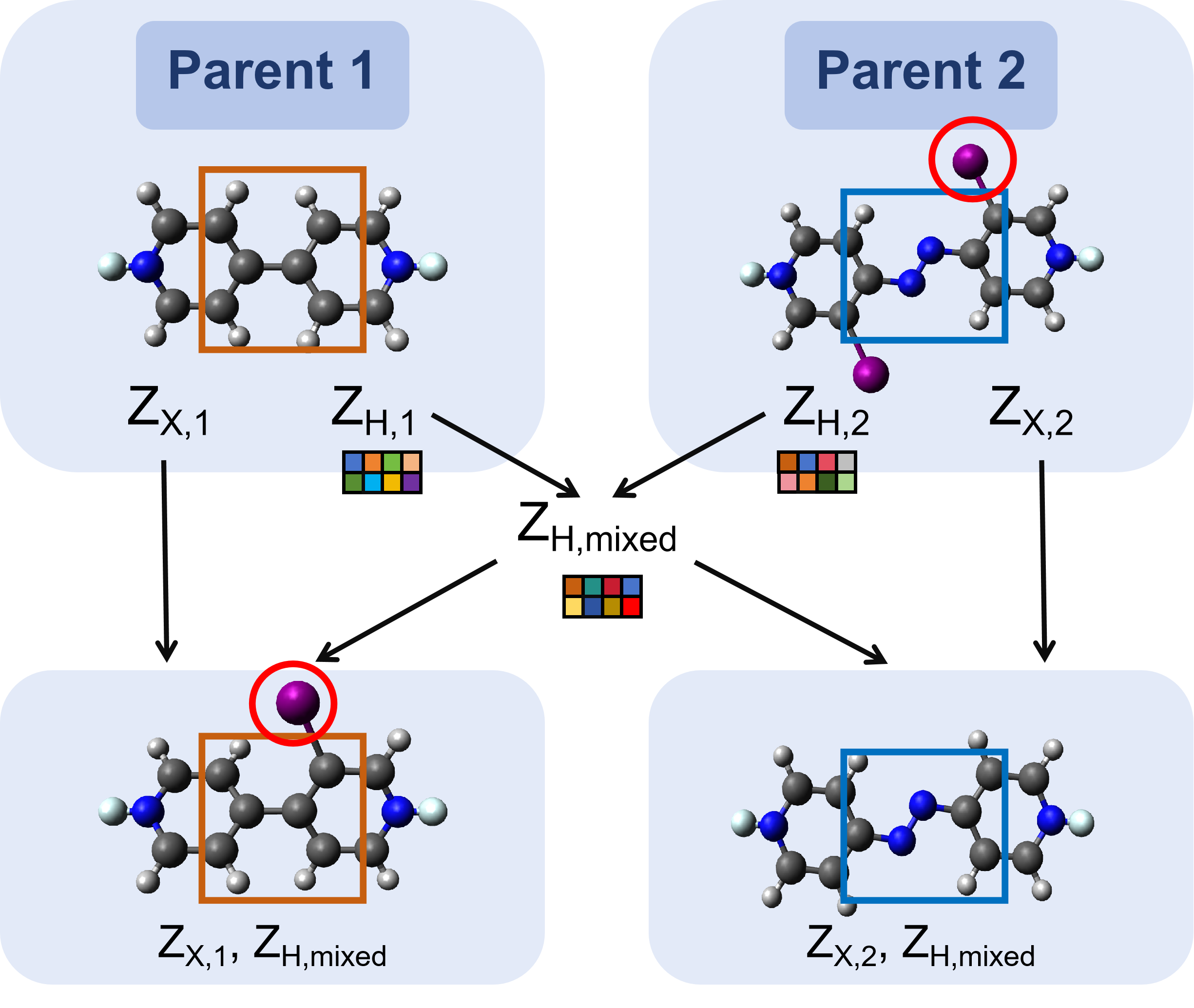}
\caption{\textbf{Latent style transfer between structurally distinct linkers.} The panel illustrates the successful disentanglement of physical geometry ($Z_X$) and chemical features ($Z_H$) within the LinkerVAE latent space. (1) \textbf{Parent 1 ($Z_{X,1}, Z_{H,1}$)}, serving as the source of the initial geometric scaffold, characterized by directly-connected carbon rings structure (highlighted by the orange box). (2) \textbf{Parent 2 ($Z_{X,2}, Z_{H,2}$)}, the original structural source of the twisted geometric backbone and the iodine substituent. (3) \textbf{Generated Linker 1 ($Z_{X,1}, Z_{H,mixed}$)}, a hybrid structure that strictly preserves the 3D spatial backbone of Parent 1 while successfully inheriting the chemical "style"— an iodine functional group (red circle)—via the interpolated chemical latent representation. (4) \textbf{Generated Linker 2 ($Z_{X,2}, Z_{H,mixed}$)}, which maintains the distinct twisted, azo-linked spatial geometry of Parent 2 (highlighted by the blue box) but exhibits an altered, averaged chemical functionalization. Terminal light-blue atoms represent the virtual anchor atoms (He) required for downstream MOF assembly.}
\label{fig:style_transfer}
\end{figure}

The results in Fig.~\ref{fig:style_transfer} and Supplementary Table 3 consistently demonstrated that LinkerVAE enables implicit functional group swapping while maintaining the desired 3D structural skeleton. Both Gen1 and Gen2 maintained 100\% chemical validity. To quantitatively assess shape preservation, we measured the Shape Tanimoto similarity (computed as $1 - \text{ShapeTanimotoDist}$), which evaluates the spatial volume overlap. Gen1 retained a high shape similarity of 0.646 with its geometric parent P1, and Gen2 achieved 0.669 with P2. Concurrently, the anchor coordinates remained tightly aligned, exhibiting mean spatial deviations of only 0.283 \AA~ and 0.253 \AA~ for Gen1 and Gen2, respectively. This orthogonal control directly supports the viability of gradient-based optimization within the latent manifold, as chemical features can be aggressively updated without catastrophically distorting the physical skeleton necessary for crystalline assembly.

\subsection{Zero-shot Isoreticular Expansion via Latent Geometric Scaling}

Having established that LinkerVAE can seamlessly decouple and independently manipulate the chemical and geometric attributes of individual linkers, we next extended this capability to the macroscopic scale of full periodic MOF structures. A key objective in materials engineering is isoreticular expansion, which involves systematically elongating organic linkers to increase the pore dimensions and surface area of MOFs while preserving their underlying crystalline topology and chemical functionality \cite{isoexpansion}. Traditionally, this requires synthesizing and screening novel extended ligands. Here, we demonstrate that our framework can achieve zero-shot isoreticular expansion entirely \textit{in silico} via targeted transformations in the continuous latent space.

For this experiment, we selected a diverse subset of existing MOFs (comprising 1 metal node and 3 distinct linkers per asymmetric unit) from a curated database of 500 MOFs. To induce expansion, we first parsed the strict rigid-body topology of each MOF, pinning down the precise Cartesian coordination vectors between the linker anchor atoms and the metal node centers under periodic boundary conditions. Next, we encoded each linker into the LinkerVAE latent space. By performing Singular Value Decomposition (SVD) on the local coordinates of the anchor atoms, we identified the principal growth axis for each linker. We then anisotropically stretched the spatial latent variable ($Z_x$) along this principal axis by a target scale factor of 1.4$\times$, while leaving the perpendicular spatial components and the chemical latent variable ($Z_h$) entirely unchanged. Concurrently, the target atom count ($N$) was proportionally scaled.

Following the decoding of these elongated linkers, we introduced a differentiable, gradient-based rigid-body assembly optimization. By initializing a proportionally expanded unit cell, we utilized an Adam optimizer to fine-tune the 3D roto-translations of the generated linkers alongside the lattice parameters. The objective function strictly minimized the Euclidean distance between the generated coordination atoms and the pinned geometric targets, ensuring that the local chemical bonding environment at the metal-organic interface was perfectly maintained.

\begin{figure}[htbp]
\centering
\includegraphics[width=1.0\textwidth]{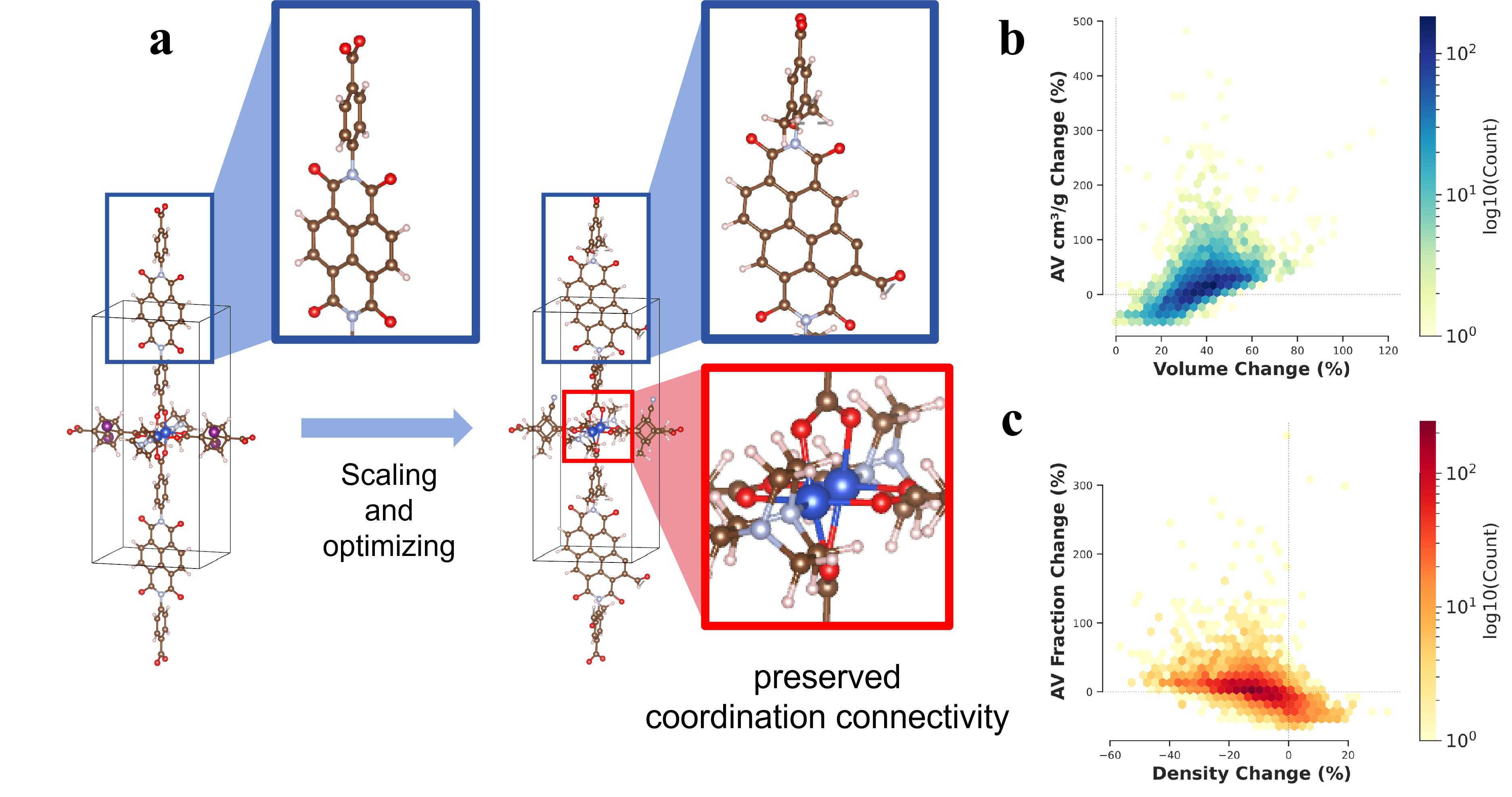} 
\caption{\textbf{Zero-shot isoreticular expansion via continuous latent scaling.} 
\textbf{a} Structural transformation of a representative MOF before (left) and after (right) the targeted 1.4$\times$ expansion process. By anisotropically stretching the spatial latent variable ($Z_x$) along the principal axes and proportionally scaling the decoded atom count, the model generates elongated linker analogues (blue inset) that seamlessly extend the structural backbone. The strict geometric optimization guarantees perfectly preserved coordination connectivity at the metal-organic interface (red inset). 
\textbf{b} Distribution of percentage changes in unit cell volume versus gravimetric accessible volume (AV, cm$^3$/g). The logarithmic point density highlights a robust positive trend where volumetric enlargement translates into significant gains in storage capacity. 
\textbf{c} Log-scaled hexbin plot illustrating the inverse correlation between framework density reduction and the enhancement of AV fraction. The high-density cluster in the upper-left quadrant confirms the consistent generation of lighter, high-porosity architectures via latent space manipulation.}
\label{fig:mof_expansion_combined}
\end{figure}

Zeo++ was used to compute porous properties of the edited MOFs \cite{zeo}. Results show that the automated isoreticular expansion was remarkably robust, yielding physically valid and chemically sound MOF structures with an 88.15\% success rate. To quantitatively evaluate the physical enhancements of the expanded materials, we computed their geometric and pore characteristics using high-accuracy geometric volume tools (Supplementary Table 4). The visualization for an example MOF being scaled and optimized is shown in Figure \ref{fig:mof_expansion_combined}a.

The latent scaling successfully induced a significant macroscopic expansion, with the average unit cell volume increasing by 38.66\% and the framework density decreasing by 11.45\%. Most notably, this geometric enlargement translated directly into enhanced porosity: the gravimetric accessible volume (AV cm$^3$/g) saw a dramatic average increase of 23.32\%, with some highly responsive frameworks exhibiting enhancements of up to 289.3\%. 

Crucially, despite these drastic volumetric changes, the difference in the number of discrete pore channels before and after expansion was exactly zero ($\Delta = 0$) across all successfully optimized MOFs. This confirms that the generative pipeline did not inadvertently collapse pores or alter the fundamental connectivity of the network. The relationship between latent structural expansion and the resulting multi-metric physical enhancements is visualized in Figure \ref{fig:mof_expansion_combined}b, c. The data reveals a clear synergy between volumetric swelling and gravimetric capacity, alongside a corresponding decrease in framework density that facilitates higher porosity. This confirms that the degree of latent geometric manipulation directly dictates the macroscopic pore characteristics of the optimized MOFs.

\subsection{Property Prediction and Test-Time Optimization via Surrogate Model}

\subsubsection{High-Fidelity Surrogate Model for MOF Properties}
To enable property-guided material discovery and optimization, we developed a fast and accurate surrogate model for predicting key MOF properties. This model employs a coarse-grained latent graph representation of the MOF structure, where each linker is condensed into a virtual node with its geometric ($Z_x$) and chemical ($Z_h$) latent features acting as node attributes, and its centroid coordinate defining the node position, while metal nodes are incorporated as global conditioning features. The model leverages a SchNet-based architecture with periodic boundary conditions (PBC) and Radial Basis Functions (RBF) for encoding inter-node distances, effectively capturing long-range interactions within the MOF framework (see Methods for details).

\begin{figure}[htbp]
\centering
\includegraphics[width=1.0\textwidth]{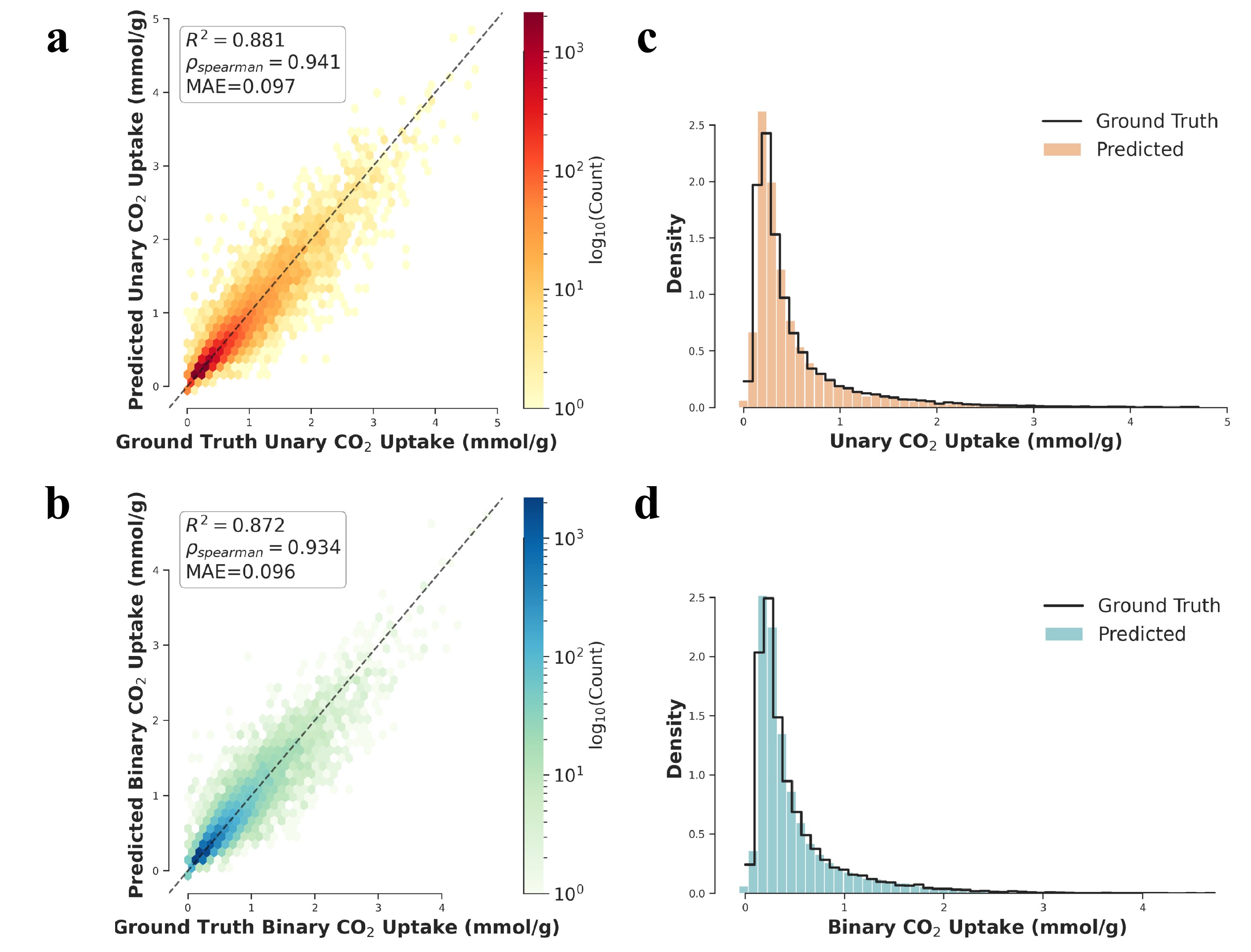}
\caption{\textbf{Surrogate model performance on MOF carbon capture properties.} (a,b) Hexbin parity plots for the validation set, comparing ground truth versus predicted CO$_2$ uptake in (a) pure CO$_2$ gas and (b) binary CO$_2$/N$_2$ gas. The density-colored points demonstrate strong correlation and predictive accuracy ($R^2$, SRCC, and MAE values are indicated). (c,d) Overlapping distribution plots showing the fidelity of the model's predictions in capturing the underlying statistical distribution across the validation set.}
\label{fig:surrogate_performance}
\end{figure}

The surrogate model was trained and evaluated to predict CO$_2$ uptake under two relevant conditions: pure CO$_2$ adsorption and CO$_2$ uptake in the presence of a CO$_2$/N$_2$ mixture (0.15 bar CO$_2$, 0.85 bar N$_2$, 298 K), in order to assess its ability to capture both absolute adsorption capacity and competitive adsorption effects. As demonstrated in Figure \ref{fig:surrogate_performance}, the model exhibits high predictive accuracy and strong rank correlation on unseen MOFs from the validation set. For pure CO$_2$ uptake, the model achieved an $R^2$ of 0.881, a Spearman's Rank Correlation Coefficient (SRCC) of 0.941, and a Mean Absolute Error (MAE) of 0.097 mmol/g. For binary CO$_2$/N$_2$ uptake, it achieved an $R^2$ of 0.872, an SRCC of 0.934, and an MAE of 0.096 mmol/g. These metrics highlight the model's robustness and capability to accurately guide property optimization.

\subsubsection{Latent Space Test-Time Optimization (TTO)}
\begin{figure}[htbp]
 \centering
 \includegraphics[width=0.95\textwidth]{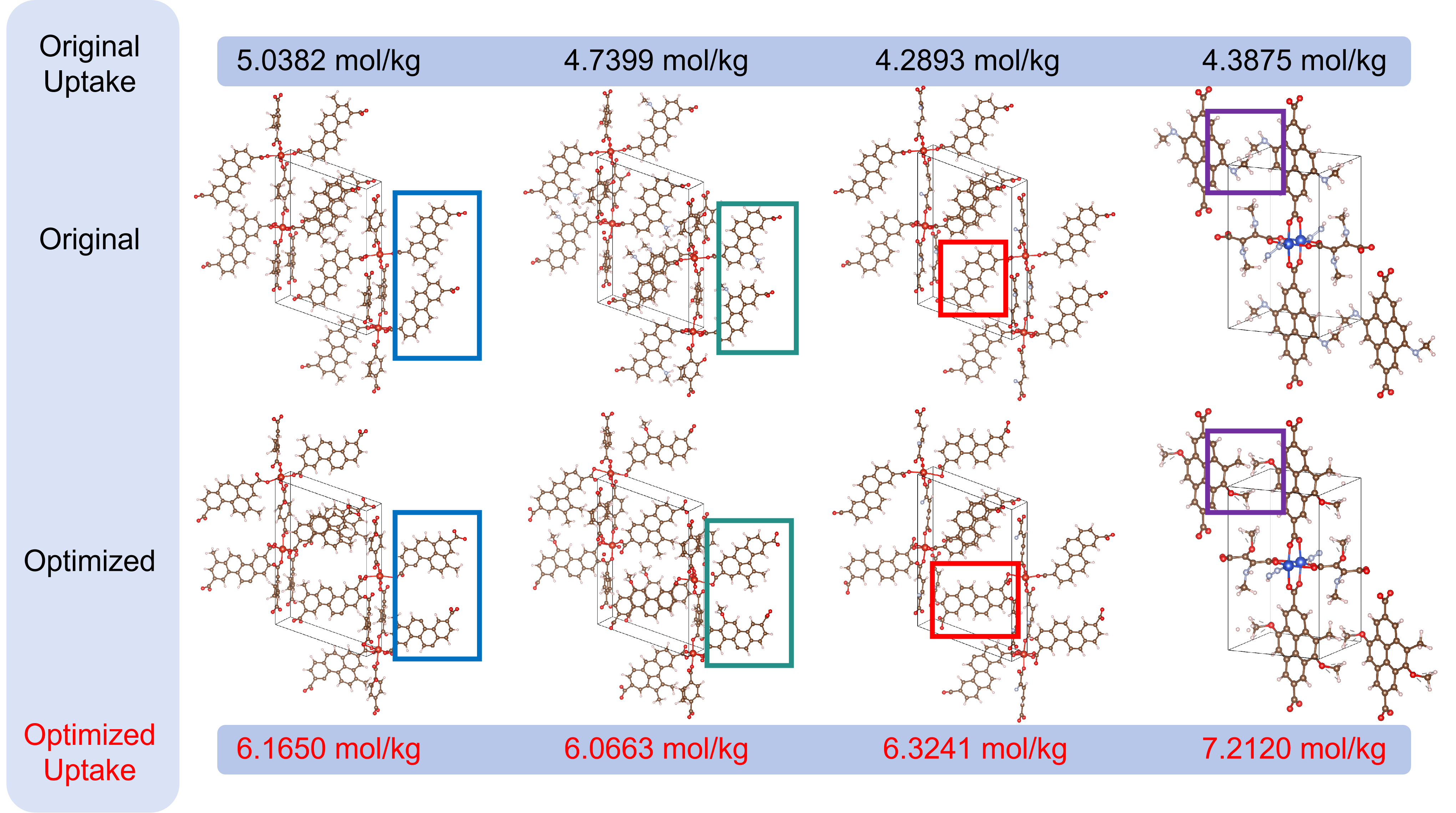} 
\caption{\textbf{Representative cases of TTO pushing the performance limits.} 
    The top row displays original MOF structures from the high-performance tail of the training set, while the bottom row shows their corresponding optimized versions.
    Colored boxes highlight the specific structural refinements identified by the model. 
    These micro-structural adjustments result in substantial increases in CO$_2$ uptake (e.g., from 4.38 to 7.21 mol/kg), demonstrating the capability of the latent manifold to suggest out-of-distribution structural motifs that extend beyond the distribution of the training dataset.}
    \label{fig:breakthrough_examples}
\end{figure}

Leveraging the differentiability of LinkerVAE's latent space and the high accuracy of our surrogate predictor, we implemented a Test-Time Optimization (TTO) strategy. This approach allows for direct, gradient-based updates to the latent variables and atom counts of an existing MOF structure, continuously evolving it towards desired properties without requiring retraining of the generative model (see Methods). Grand Canonical Monte Carlo (GCMC) simulations were performed using RASPA2 \cite{raspa} to evaluate the adsorption property changes induced by the optimization. We performed three distinct TTO experiments:

\begin{enumerate}
    \item \textbf{General Optimization for Pure CO$_2$ Uptake (Exp. 1):} We applied TTO to a diverse set of 500+ MOFs, optimizing for pure CO$_2$ uptake. As detailed in Supplementary Table 5 and Figure \ref{fig:tto_results}a, 74.0\% of the optimized structures exhibited improved CO$_2$ uptake, with a substantial mean relative boost of 147.5\% and a median relative boost of 23.6\%. 
    \item \textbf{General Optimization for CO$_2$ Uptake in Binary CO$_2$ (0.15 bar) / N$_2$ (0.85 bar) Condition (Exp. 2):} Similar to Exp. 1, TTO was applied to 500+ MOFs to maximize CO$_2$ uptake in mixed flue gas. We observed that 70.3\% of the MOFs improved their performance, achieving a mean relative boost of 94.8\% and a median boost of 13.7\% (Supplementary Table 5, Figure \ref{fig:tto_results}b). This confirms the TTO's effectiveness across different conditions.
    \item \textbf{High-Performance Ceiling Breakthrough (Exp. 3):} To push the boundaries of materials discovery, we then applied TTO to a selection of 1000 already top-performing MOFs, constituting the highest 0.33\% of the original dataset for pure CO$_2$ uptake. Even for these exceptionally performing materials, 44.9\% showed further improvements, and TTO successfully broke the performance ceiling of the initial dataset (Fig.~\ref{fig:breakthrough_examples}). As illustrated, the optimization process does not merely perform random perturbations; instead, it executes targeted structural refinements within the continuous latent manifold.
\end{enumerate}

These results collectively highlight the power of our differentiable latent space combined with surrogate-guided TTO, providing a targeted and efficient pathway for property optimization and novel material discovery.

\begin{figure}[htbp]
\centering
\includegraphics[width=1.0\textwidth]{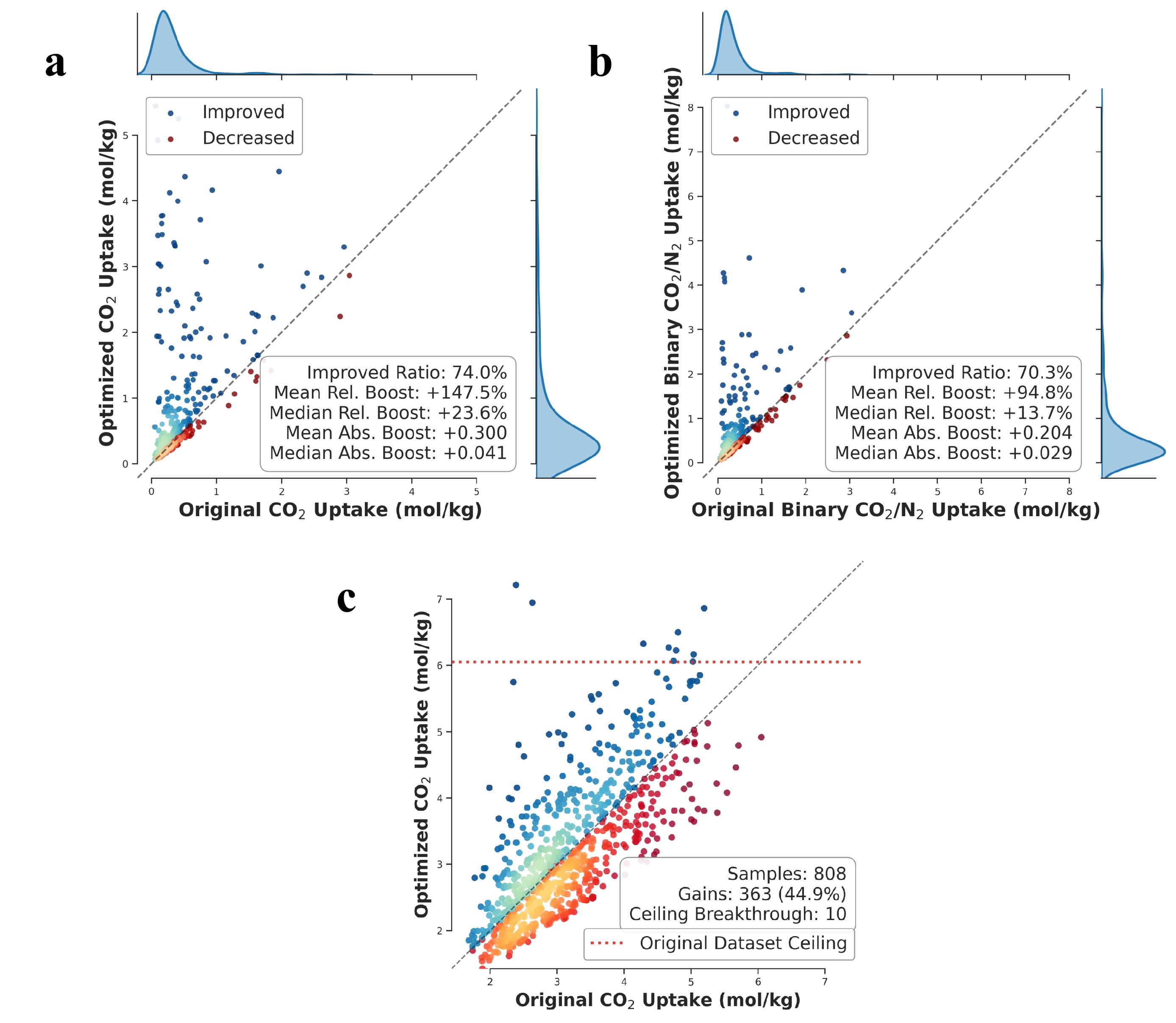}
\caption{\textbf{Test-Time Optimization (TTO) results for MOF property enhancement.} (a,b) Categorized density scatter plots for general TTO, showing original versus optimized property values. Points are colored by density for improved (blue-green) and decreased (red-orange) performance. (a) Optimization for pure CO$_2$ uptake (Exp. 1). (b) Optimization for binary CO$_2$/N$_2$ condition (Exp. 2). Statistical improvements are detailed in Supplementary Table 5. (c) Ceiling breakthrough analysis for pure CO$_2$ uptake (Exp. 3), focusing on the top 0.33\% of original MOFs (1000 samples). The dashed line represents the original dataset's maximum performance, which the TTO process successfully surpassed.}
\label{fig:tto_results}
\end{figure}

\subsubsection{Latent Manifold Topology and Interpretability}
To gain deeper insights into the interpretability and organization of our latent space, we performed a Uniform Manifold Approximation and Projection (UMAP \cite{umap}) and Hierarchical Density-Based Spatial Clustering of Applications with Noise (HDBSCAN \cite{HDBSCAN}) analysis on the 256-dimensional global latent features ($h_{global}$) extracted by the surrogate model for pure CO$_2$ uptake.

\begin{figure}[htbp]
\centering
\includegraphics[width=1.0\textwidth]{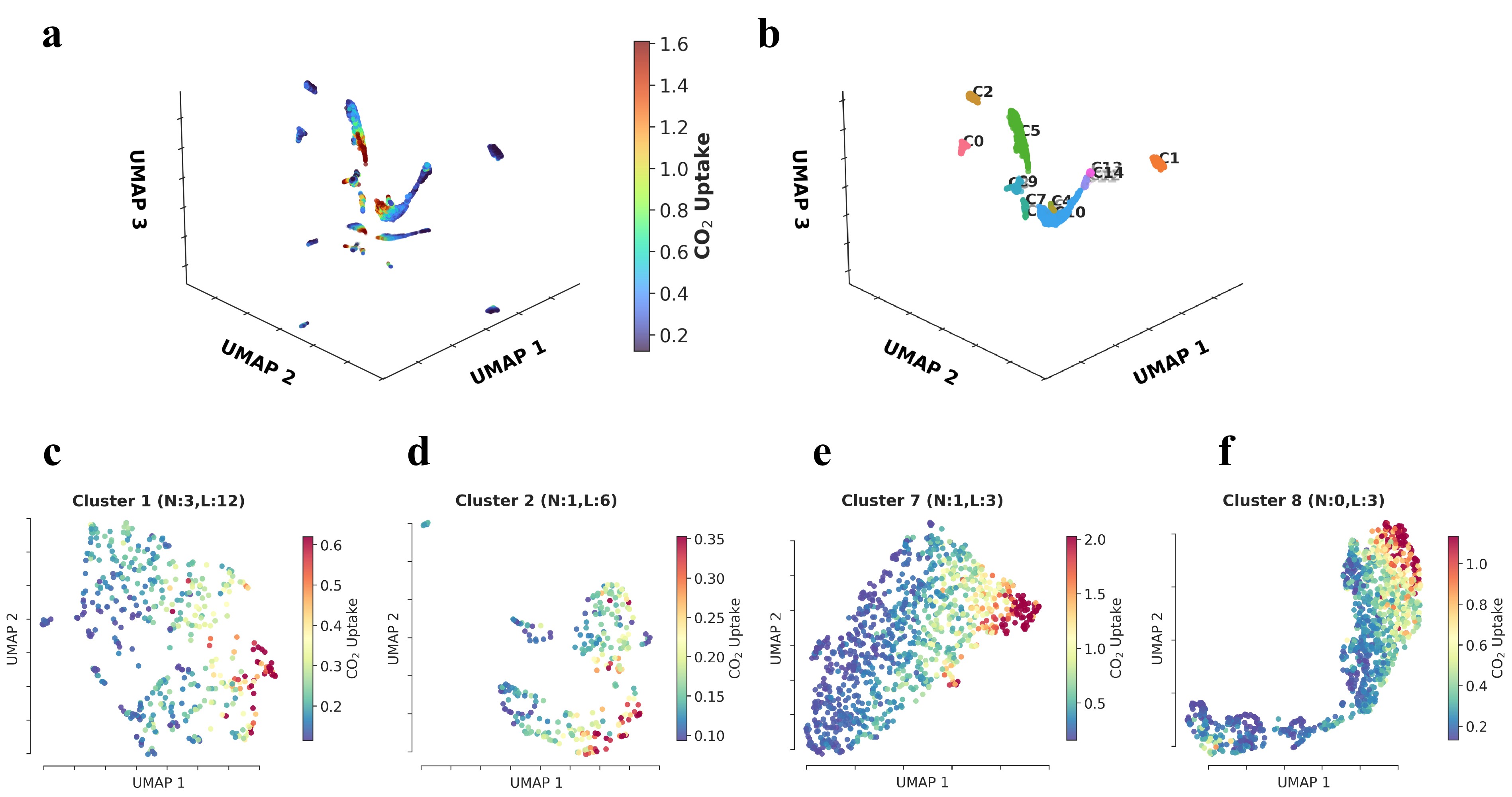} 
\caption{\textbf{Topology and interpretability of the global latent manifold.} UMAP dimensionality reduction of the 256-dimensional global latent features ($h_{global}$) extracted from the surrogate model for pure CO$_2$ uptake. (a) A 3D projection of the entire latent manifold, colored by CO$_2$ uptake, revealing distinct clusters and a clear property gradient. (b) The same 3D manifold, with points colored by their HDBSCAN cluster assignments. Each cluster is annotated with its dominant topological profile (Node connectivity:Linker count), demonstrating that the latent space naturally organizes MOFs into distinct topological families. (c–f) 2D UMAP projections for four representative clusters. Within each topologically homogeneous cluster, the latent space is smooth, exhibiting clear and consistent property gradients (CO$_2$ uptake) that enable robust targeted optimization.}
\label{fig:latent_manifold}
\end{figure}

As depicted in Figure \ref{fig:latent_manifold}a, the UMAP projections reveal a highly structured latent manifold, characterized by distinct "islands" or clusters rather than a continuous, amorphous blob. This inherent discontinuity in the global latent space suggests a natural segregation of MOF structures into topologically distinct families. Upon applying HDBSCAN clustering (Figure \ref{fig:latent_manifold}b), we found that each identified cluster predominantly corresponds to MOFs sharing similar topological profiles, such as the number of linkers coordinating to the metal node and the overall linker count. 

Further examination of the intra-cluster manifolds (Figure \ref{fig:latent_manifold}c–f) demonstrates remarkable clarity. Within each topological family, the latent space is smooth, with points uniformly distributed and clear property gradients (e.g., CO$_2$ uptake levels indicated by color) highlighting specific directions of improvement or degradation. This smooth and interpretable local manifold is critical, as it underpins the ability of our TTO strategy to perform targeted, chemically valid optimization without generating anomalous or topologically inconsistent structures.

\subsection{\textit{De Novo} Linker Design via Equivariant Latent Diffusion}

While Test-Time Optimization (TTO) effectively refines existing MOFs, the discovery of entirely unprecedented framework architectures necessitates sampling fundamentally new building blocks. To achieve this, we developed a Linker Latent Diffusion Model (LLDM). By training a continuous-time equivariant diffusion model directly on the information-dense $(Z_x, Z_h)$ latent space, we can unconditionally generate novel chemical and geometric latent pairs. 

To map these generated continuous latents back into discrete 3D molecular graphs, our decoding pipeline requires the target atom count ($N$). Rather than treating $N$ as a random variable, we trained an auxiliary invariant Graph Neural Network (GNN) regressor to predict $N$ directly from the generated latent representations. Evaluated on a hold-out validation set of 151,115 samples, the regressor demonstrated exceptional precision, achieving a Mean Absolute Error (MAE) of only 0.534 atoms. Crucially, 67.25\% of predictions were exactly correct, and 92.74\% fell within a strict tolerance of $\pm 1$ atom. This highly accurate mapping ensures that the subsequently generated atom counts remain consistent with the spatial and chemical capacities of BFN decoding process.

To evaluate the generative capacity of the LLDM, we unconditionally sampled 1,000 \textit{de novo} latent vectors, predicted their corresponding atom counts, and decoded them into explicit 3D molecular graphs. We assessed the generated linkers using the standard metrics of Validity (V), Uniqueness (U), and Novelty (N) relative to the training dataset containing over 275,000 unique linkers. As summarized in Supplementary Table 6, our framework achieved an outstanding chemical validity of 97.50\%. Among the valid structures, 79.69\% were unique, and of those, 82.11\% were completely novel (absent from the training set), yielding an absolute yield of 63.80\% for molecules that are both unique and novel.

\begin{figure}[htbp]
\centering
\begin{subfigure}{0.45\textwidth}
\includegraphics[width=\textwidth]{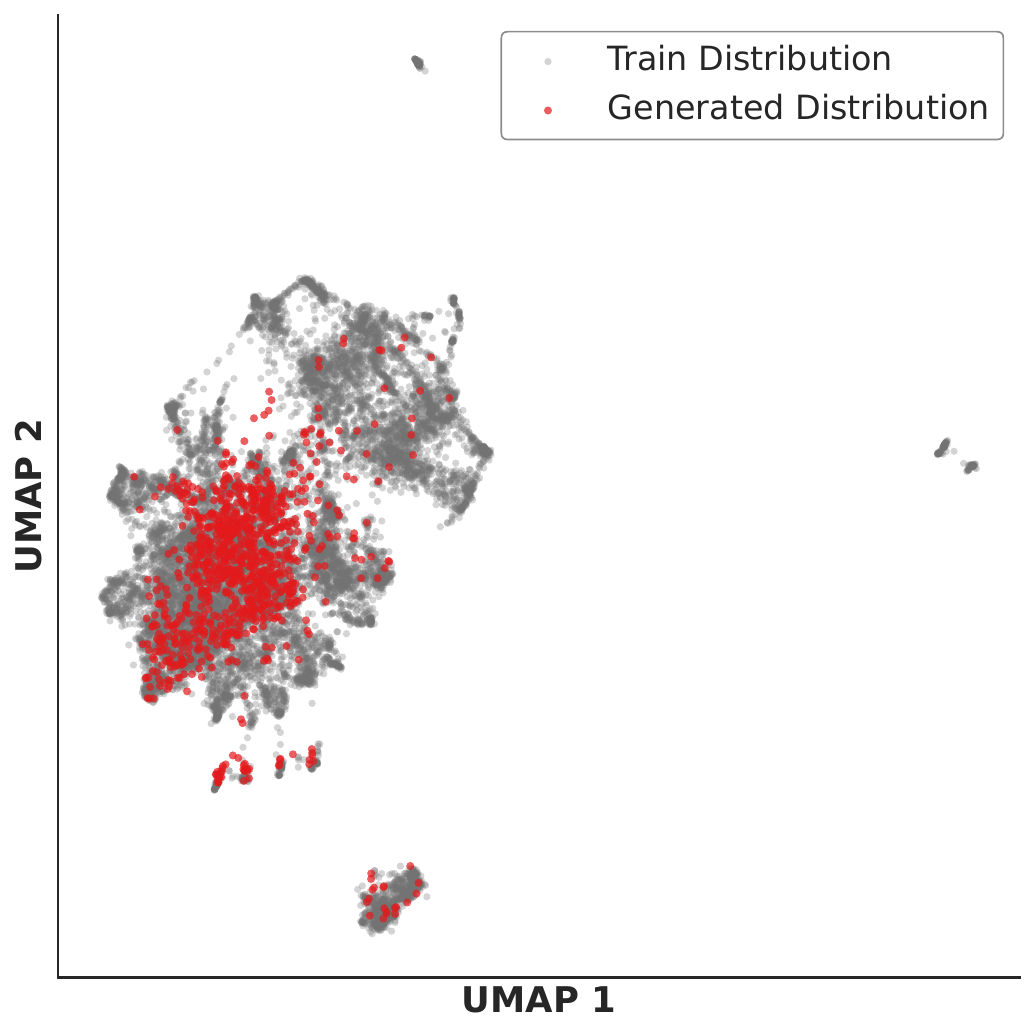}
\caption{Latent Space UMAP Overlap}
\end{subfigure}
\hfill
\begin{subfigure}{0.45\textwidth}
     \includegraphics[width=\textwidth]{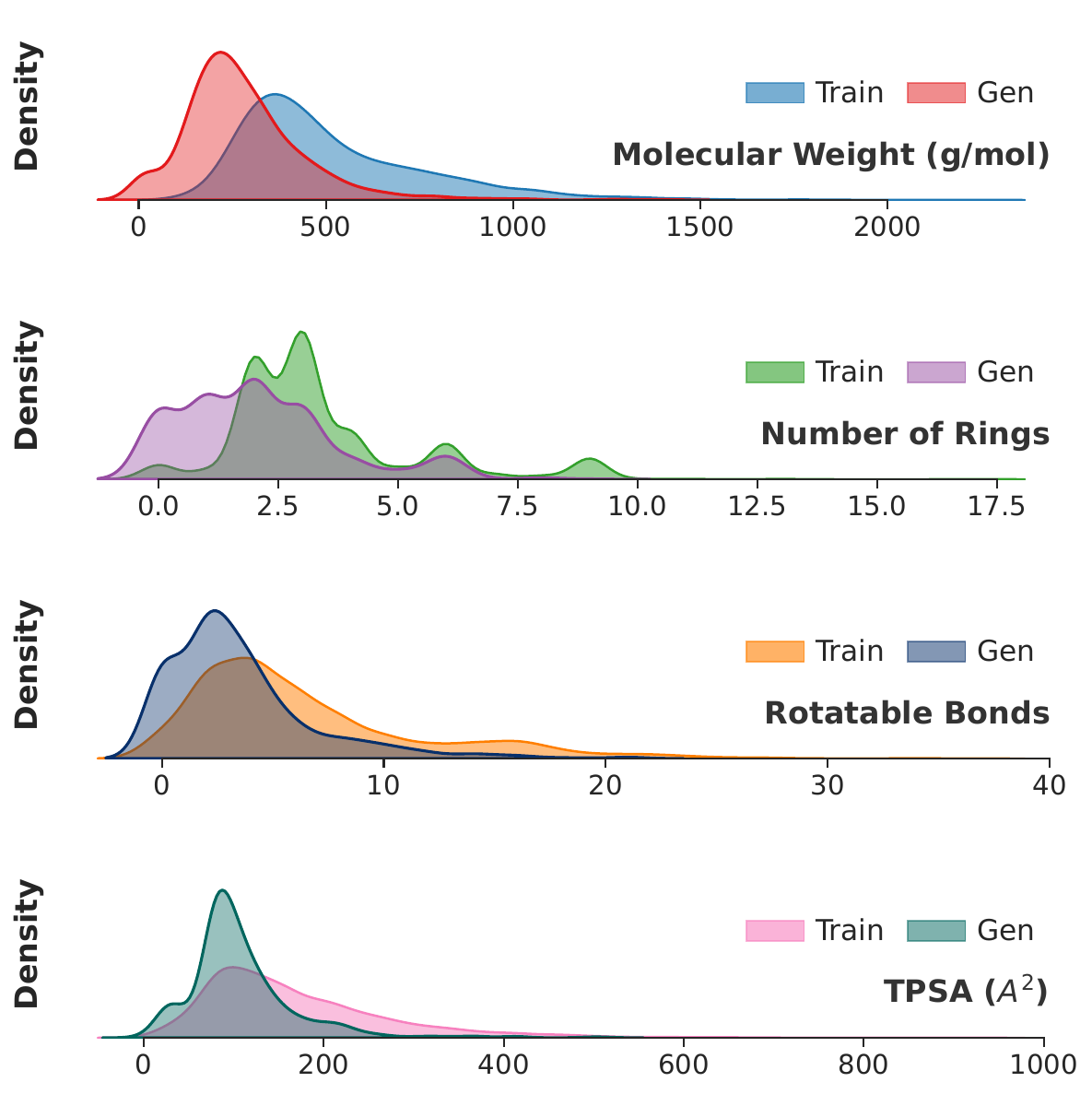}
     \caption{Chemical Descriptor Distributions}
\end{subfigure}
\caption{\textbf{Distributional alignment of LLDM-generated linkers.} (a) 2D UMAP projection comparing the continuous $(Z_x, Z_h)$ latent space of the original training set (gray) against the unconditionally generated latents (red). The high degree of overlap indicates that the LLDM successfully captures the true data distribution without mode collapse. (b) Density distributions of four key chemical descriptors (Molecular Weight, Number of Rings, Rotatable Bonds, and Topological Polar Surface Area) comparing the training dataset against the valid generated molecules.}
\label{fig:lldm_distribution}
\end{figure}

Beyond basic validity, it is imperative that the generated molecules inhabit a chemically realistic parameter space. We performed UMAP analysis on the latent space, which confirmed that the generated latents exhibit a distribution that broadly and smoothly aligns with the original training manifold (Figure \ref{fig:lldm_distribution}a). Furthermore, we calculated key macroscopic chemical descriptors—Molecular Weight (MW), Number of Rings, Rotatable Bonds, and Topological Polar Surface Area (TPSA)—for both the generated and training sets. As shown in Figure \ref{fig:lldm_distribution}b, the property distributions of the LLDM-generated novel linkers are within the range of the true dataset. This structural and chemical alignment proves that the model learns the underlying physical design rules of MOF building blocks rather than memorizing specific patterns.

\begin{figure}[htbp]
 \centering
 \includegraphics[width=0.9\textwidth]{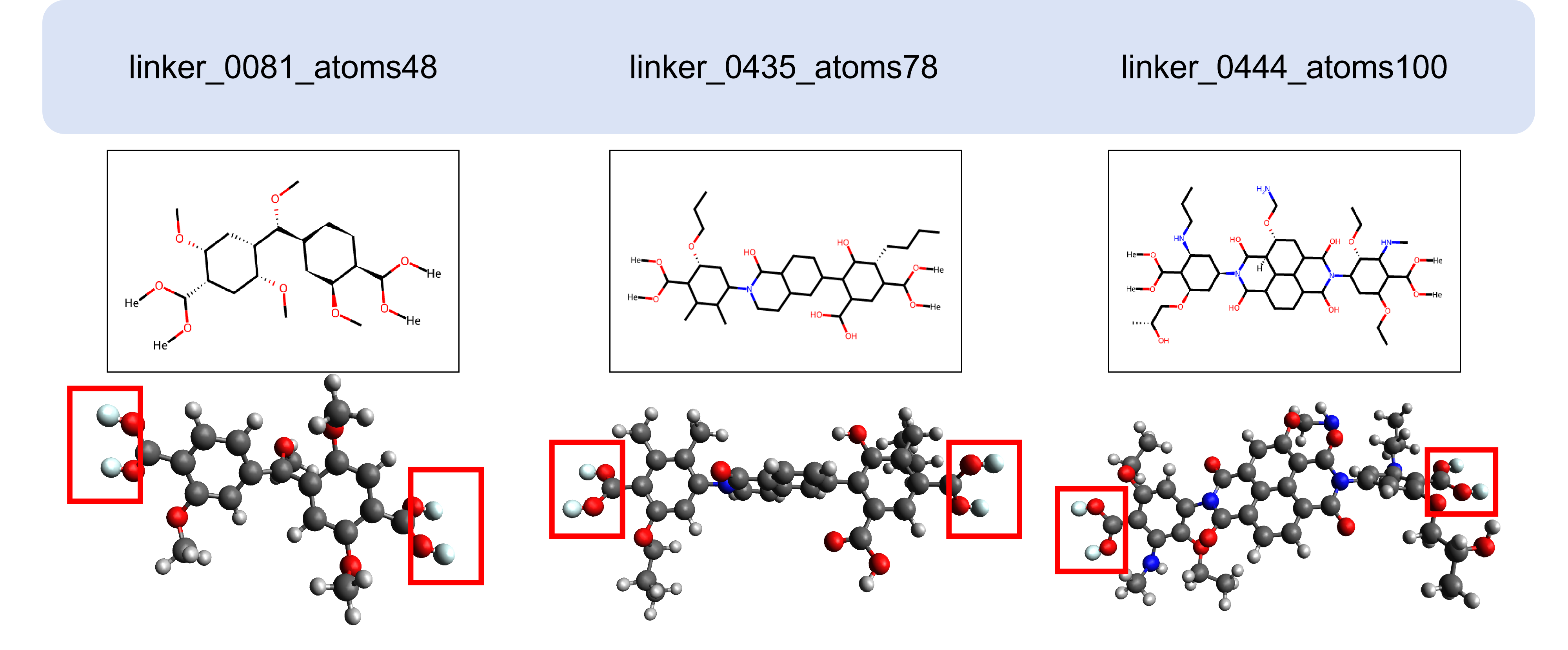} 
\caption{\textbf{Representative \textit{de novo} generated organic linkers.} Three uniquely generated and novel linker molecules produced unconditionally by the LLDM. For each sample, both the decoded 3D point cloud configurations (bottom) and their corresponding 2D chemical graphs (top) are displayed. The model demonstrates the capability to generate sophisticated topological features, including complex fused ring systems and diverse functional groups, while maintaining the precise coordination geometries required for downstream MOF assembly.}
 \label{fig:novel_linkers_examples}
 \end{figure}

The true success of the LLDM generative pipeline is visually evident in the quality of the individual structures. Figure \ref{fig:novel_linkers_examples} showcases three representative generated complex linkers that are unique and novel. The model successfully synthesizes highly complex, multi-ring aromatic systems and non-trivial heteroatom functionalizations, while perfectly preserving the critical spatial trajectories of the terminal coordination anchors necessary for rigid-body lattice assembly. These generated molecules are fully prepared to be integrated into traditional assembly methods \cite{tobacco, pormake} or modern flow-matching-based assembly modules \cite{mofflow} for the construction of new MOF architectures.

\section{Discussion}\label{sec12}

The design of metal–organic frameworks (MOFs) is fundamentally limited by the enormous configurational space of atomic arrangements and the intricate chemical constraints governing their assembly. While recent deep generative models have made strides in \textit{de novo} generation, they overwhelmingly operate as ``one-shot'' black boxes. This inherently precludes the post-generation refinement and continuous structural evolution required for practical materials engineering. Our proposed LEGO-MOF framework fundamentally shifts this paradigm, bridging the gap between machine-learning-driven molecular generation and the rigorous geometric rules of reticular chemistry. By embedding variable-length, 3D chemical graphs into a fully differentiable, SE(3)-equivariant latent manifold, we demonstrate that complex MOF architectures can be manipulated, scaled, and optimized with the same fluidity as continuous numerical variables.

The foundation of this framework lies in the strict disentanglement of spatial geometry ($Z_x$) and invariant chemical features ($Z_h$) within the LinkerVAE latent space. In traditional molecular representations, minor structural perturbations often lead to catastrophic failures in 3D geometry or invalid chemical valency \cite{jtvae, kusner2017grammar, polykovskiy2020moses}. By contrast, our approach guarantees that the spatial scaffold remains highly regularized. A key design choice in our model is the explicit incorporation of virtual “anchor” atoms (helium), which are introduced to guide the model in recognizing coordination sites and to enforce the chemical consistency between linkers and the resulting MOF structures. In reticular chemistry, the predictable directionality of coordination bonds is paramount. Our results demonstrate that even under aggressive latent perturbations—such as forcefully shifting the heavy atom count (Supplementary Table 2) or swapping entire functional groups via style transfer (Fig.~\ref{fig:style_transfer})—the model steadfastly maintains the spatial trajectories of these anchors. This transforms abstract latent vectors into physically viable structural modules that are intrinsically prepared for downstream rigid-body lattice assembly, circumventing the geometric collapse that plagues many existing 3D generative pipelines.

Building upon this disentangled manifold, LEGO-MOF unlocks zero-shot \textit{in silico} manipulations that closely mirror physical reticular chemistry. The zero-shot isoreticular expansion achieved via anisotropic latent geometric scaling represents a powerful capability. Traditionally, achieving isoreticular expansion requires synthesizing novel, elongated organic ligands—a process fraught with solubility and stability issues. Our framework executes this entirely computationally by stretching $Z_x$ along its principal growth axis, dynamically adjusting the atom count and post-optimizing. The strict quantitative relationship observed between the degree of latent scaling and macroscopic properties—namely, a dramatic 23.32\% average increase in gravimetric accessible volume without disrupting the topological channel count (Fig.~\ref{fig:mof_expansion_combined}a, b)—provides compelling evidence that the latent space faithfully encodes the underlying physical principles of volumetric swelling and porosity, rather than merely interpolating surface-level visual patterns.

Another impactful application of this continuous manifold is the surrogate-guided Test-Time Optimization (TTO). In typical materials discovery, researchers are confined to the interpolation bounds of their training datasets. TTO bypasses this limitation by treating the pre-trained generative model and the surrogate property predictor as a unified differentiable system. Instead of representing MOFs as  unified information-incomplete latent vectors \cite{mofdiff, xrd2mof}, we encode them as coarse-grained latent graphs that preserve both spatial and chemical information while achieving substantial dimensionality reduction. As demonstrated by the UMAP and HDBSCAN clustering analyses (Fig.~\ref{fig:latent_manifold}), the global latent space naturally self-organizes into distinct topological families. Within these topological ``islands,'' the local manifold is remarkably smooth, exhibiting consistent, differentiable property gradients. This intra-cluster smoothness is the exact mechanism that allows TTO to perform stable gradient ascent. By traversing these local gradients, TTO dynamically sculpts existing structures to maximize CO$_2$ uptake without altering the fundamental framework topology. Most notably, applying TTO to the top 0.33\% of high-performing MOFs successfully broke the performance ceiling of the original dataset (Fig.~\ref{fig:breakthrough_examples}). The emergence of micro-structural refinements that yield substantial performance boosts demonstrates that the model is actively discovering out-of-distribution structural motifs, transitioning from a passive data-interpolator to an active materials optimizer.

To complete the materials discovery pipeline, the integration of the Linker Latent Diffusion Model (LLDM) expands the framework's reach into completely uncharted chemical spaces. The exceptionally high absolute yield of valid, unique, and novel linkers (63.80\%, as detailed in Supplementary Table 6) confirms that the LLDM learns the generative physical rules of organic fragments rather than memorizing the training set. When coupled with flow-matching-based or traditional topological assembly algorithms, LEGO-MOF establishes a fully automated, end-to-end pipeline: from the continuous generation of novel linkers to the targeted optimization of assembled crystal properties.

Despite these significant advancements, several limitations present opportunities for future exploration. First, although the generated structures achieve near-perfect chemical validity, \textit{validity} does not inherently guarantee \textit{synthesizability}. Future iterations of the model would benefit from incorporating synthetic accessibility (SA) scores or reaction-template constraints directly into the latent optimization objective. Second, MOFs are inherently dynamic materials, often exhibiting structural flexibility, or ``breathing'' effects, upon gas adsorption. Traditional assembly methods or recent flow-matching assembly pipeline assumes rigid-body mechanics. Incorporating molecular dynamics (MD) or flexible force-field relaxations into the differentiable loop could yield more accurate property predictions for flexible frameworks.

In conclusion, LEGO-MOF establishes a highly modular, target-driven paradigm for the design of metal-organic frameworks. By translating the discrete, combinatorial problem of reticular chemistry into a continuous, physics-aware optimization problem within an SE(3)-equivariant latent space, we provide a robust computational pathway for the targeted editing and enhancement of advanced functional materials. We anticipate that this framework will seamlessly integrate with autonomous robotic laboratories (self-driving labs), where continuous latent optimization can be directly coupled with high-throughput synthesis and characterization, accelerating the deployment of next-generation porous materials for global energy and environmental challenges.

\section{Methods}\label{sec11}

\subsection{Data Source and Building Block Decomposition}

The primary data source for this study is the BW-DB dataset, a collection of approximately 304k hypothetical Metal-Organic Frameworks (MOFs) with less than 20 building
blocks \cite{bwdb}. The dataset was precessed by MOFDiff \cite{mofdiff} to structurally decomposite MOFs into discrete building blocks using the "metal-oxo" algorithm provided by the MOFID toolkit \cite{mofid}.

Specifically, the decomposition algorithm identifies and severs the coordination bonds between the central metal atoms and the coordinating oxygen or nitrogen atoms of the organic ligands. To preserve the topological information and coordination geometry required for downstream framework re-assembly, virtual anchor atoms (Helium) were inserted at each cleavage site on the organic fragments. This process transformed the periodic MOF crystals into a library of approximately 1.51 million discrete organic linkers.

The extracted linker library was randomly partitioned into a training set (95\%) and a validation set (5\%) using a fixed manual seed of 42 to ensure reproducibility. The training samples were processed and stored using a memory-mapped database (LMDB) to ensure high-performance data loading. During the pre-processing phase, atomic species were mapped to a discrete vocabulary of 17 atom types, and the structural metadata were serialized to synchronize the feature dimensions across the LinkerVAE, surrogate, and latent diffusion modules.

\subsection{LinkerVAE Architecture and Training}

\textbf{Equivariant Encoding.} The encoder utilizes an equivariant Transformer architecture (9 layers, 128 hidden dimensions, 16 attention heads) to process variable-length linker point clouds. Through a set of 10 virtual global nodes, the encoder compresses the structural information into a disentangled latent representation $Z = (Z_x, Z_h)$. $Z_h \in \mathbb{R}^{32}$ captures invariant chemical and topological features, while $Z_x \in \mathbb{R}^{10 \times 3}$ encodes the equivariant spatial scaffolding. 

\textbf{Bayesian Flow Decoding.} The decoder is formulated as a continuous-time Bayesian Flow Network (BFN), which models the joint distribution of atomic coordinates and chemical identities. During training, the model learns to reconstruct the original linker from the latent variables by minimizing the Bayesian flow loss over $t \in [0, 1]$. To ensure high fidelity at coordination sites—critical for valid MOF assembly—we introduced an \textit{anchor-weighting} scheme, applying a $1.5\times$ penalty factor to the reconstruction loss for the virtual Helium anchor atoms ($Z=2$). 

\textbf{Objective Function and Optimization.} The total training objective is a weighted combination of the BFN reconstruction loss ($\mathcal{L}_{recon}$) and the Kullback-Leibler (KL) divergence ($\mathcal{L}_{KL}$):
\begin{equation}
    \mathcal{L}_{total} = w_{recon}(w_c \mathcal{L}_{coord} + w_d \mathcal{L}_{type}) + w_{kl}(w_{zh} \mathcal{L}_{zh} + w_{zx} \mathcal{L}_{zx})
\end{equation}
where $\mathcal{L}_{coord}$ and $\mathcal{L}_{type}$ represent the continuous coordinate and discrete atom-type losses, respectively. We set $w_{c}=2.0$ and $w_{d}=1.0$ to prioritize geometric accuracy. The KL term employs a standard Gaussian prior for $Z_h$ and a broadened Gaussian prior ($\sigma^2=10.0$) for $Z_x$ to accommodate larger geometric variances. The model was trained using the Adam optimizer ($LR=3 \times 10^{-4}$, $\beta_1=0.95, \beta_2=0.999$) with a plateau-based learning rate scheduler (factor 0.6, patience 10) for 30 epochs, utilizing a batch size of 64 and a gradient clipping threshold of 1.0.

\subsection{Latent Space Structural and Chemical Editing}

To evaluate the disentanglement and steerability of the LinkerVAE manifold, we performed three targeted editing experiments, as described in Section~\ref{sec2}.

\textbf{Atomic Number Editing.} We assessed the decoder's sensitivity to discrete atom count conditions ($N_{atoms}$) by encoding a linker into $(Z_x, Z_h)$ and subsequently decoding it with a modified atom count $N_{edit} = N_{orig} + \Delta$, where $\Delta \in \{-3, \dots, +3\}$. For each perturbation, we reconstructed the 3D molecular graph using a bond-order-free builder and calculated chemical validity via RDKit. The preservation of the coordination environment was quantified by measuring the spatial deviation of the decoded virtual anchor atoms relative to their original positions.

\textbf{Chemical Style Transfer.} This experiment aimed to transplant functional characteristics between structurally similar linkers. Candidate linkers with similar anchor distances were first centered and rotationally aligned such that their primary He--He coordination axis coincided with the Cartesian $Z$-axis. Given two linkers $L_1$ and $L_2$, we performed linear interpolation in the chemical latent space to obtain $Z_{h,mix} = 0.5 Z_{h,1} + 0.5 Z_{h,2}$. Hybrid linkers were then generated by combining the mixed style $Z_{h,mix}$ with the geometric scaffold $Z_{x,1}$ or $Z_{x,2}$. We utilized MACCS keys for topological similarity and Shape Tanimoto overlap for 3D geometric consistency to verify the successful decoupling of "style" from "scaffold."

\textbf{Isoreticular Expansion via Latent Scaling.} To achieve macroscopic MOF expansion, we performed anisotropic latent scaling. For a target linker, we identified its principal growth axis ($v_1$) via Singular Value Decomposition (SVD) of the local anchor atom coordinates. The geometric latent variable $Z_x$ was stretched along $v_1$ by a factor of 1.4, while the perpendicular components remained unchanged. To re-assemble the expanded MOF, we utilized a gradient-based optimization pipeline (Adam optimizer, 400 steps) that simultaneously adjusted the unit cell parameters and the 3D roto-translations of the generated linkers to minimize the Euclidean distance to the rigid-body coordination targets pinned in the crystalline framework.

\subsection{Surrogate Property Prediction and TTO}
\subsubsection{Surrogate Model}

To guide the optimization process, we developed a high-fidelity surrogate model capable of predicting gas uptake properties directly from the latent representations of MOF building blocks. 

\textbf{Coarse-Grained Crystal Graph.} We represent each MOF as a coarse-grained graph where each linker is condensed into a single node. The initial features for these nodes are the concatenated LinkerVAE latent variables $(Z_h, Z_x)$ and the atom count $N_{atoms}$. Metal nodes are treated as global categorical conditions, mapped to a continuous embedding space $\mathbb{R}^{32}$.

\textbf{Model Architecture.} The surrogate utilizes a SchNet-based backbone \cite{schnet} modified for periodic systems. Inter-linker distances $d_{ij}$ within a 35 \AA~cutoff are calculated under periodic boundary conditions (PBC) and expanded using a Radial Basis Function (RBF) layer:
\begin{equation}
    e_{ij} = \exp\left(-\gamma \left\| d_{ij} - \mu_k \right\|^2\right)
\end{equation}
The message-passing blocks employ Feature-wise Linear Modulation (FiLM) to integrate the metal node embeddings into the linker features. After 6 layers of equivariant interaction, a global attention pooling layer aggregates linker embeddings into a framework-level representation, followed by a 3-layer MLP to regress the property $y$.

\textbf{Training Protocol.} The surrogate was trained to predict $\text{CO}_2$ uptake in both pure and binary ($\text{CO}_2/\text{N}_2$) environments. We utilized a Mean Squared Error (MSE) loss defined in the Z-score normalized property space:
\begin{equation}
    \mathcal{L}_{MSE} = \frac{1}{B} \sum_{i=1}^{B} \left( \hat{y}_i - \frac{y_i - \mu_y}{\sigma_y} \right)^2
\end{equation}
Optimization was performed using AdamW ($LR=7 \times 10^{-4}$, weight decay $10^{-4}$) for 30 epochs with a batch size of 1024.

\subsubsection{Latent Test-Time Optimization (TTO)}

We leverage the differentiability of the surrogate model $f_{\text{surr}}$ and the LinkerVAE manifold to perform targeted property optimization of existing MOF structures.

\textbf{Optimization Objective.} For a given MOF, we freeze the metal nodes and the spatial scaffolding $Z_x$ to preserve the underlying topology, while treating the chemical latent features $Z_h$ and the atom count $N_{atoms}$ as continuous optimization variables. The TTO objective function is defined as:
\begin{equation}
    \mathcal{L}_{TTO} = -f_{\text{surr}}(Z_h, Z_x, N) + \lambda_{zh} \left\| Z_h - Z_{h,init} \right\|^2 + \lambda_{n} \cdot \text{Penalty}(N)
\end{equation}
where $Z_{h,init}$ is the original encoding of the linker. The first term drives the structure toward higher gas uptake, the second term acts as a manifold regularizer to prevent the latent vector from drifting into unphysical regions, and the third term constrains the atom count within a valid range $[0.5N, 2.0N]$. We perform gradient ascent using the Adam optimizer ($LR=0.005$) for 250 steps. Upon convergence, the optimized $Z_h^*$ and $N^*$ are passed to the LinkerVAE BFN decoder to generate the refined organic linkers, which are then re-assembled into the final optimized structure.

\subsection{Equivariant Linker Latent Diffusion Model (LLDM)}

To explore novel chemical spaces beyond the training distribution, we developed an equivariant Linker Latent Diffusion Model (LLDM) that operates within the continuous manifold established by LinkerVAE.

\textbf{Generative Diffusion Process.} We define the joint latent state as $\mathbf{z} = [\mathbf{z}_x, \mathbf{z}_h] \in \mathbb{R}^{10 \times (3+32)}$. The forward process $q(\mathbf{z}_t | \mathbf{z}_0)$ gradually adds Gaussian noise to the latent variables according to a polynomial schedule:
\begin{equation}
    \mathbf{z}_t = \alpha_t \mathbf{z}_0 + \sigma_t \boldsymbol{\epsilon}, \quad \boldsymbol{\epsilon} \sim \mathcal{N}(\mathbf{0}, \mathbf{I})
\end{equation}
where $\alpha_t$ and $\sigma_t$ are defined by the noise schedule $\gamma(t)$ such that $\alpha_t^2 = \text{sigmoid}(-\gamma(t))$. To maintain SE(3) equivariance, the center of mass for the geometric component $\mathbf{z}_x$ is strictly constrained to zero throughout the diffusion process.

The reverse denoising process $p_\theta(\mathbf{z}_{s} | \mathbf{z}_t)$ is parameterized by an Equivariant Graph Neural Network (EGNN) dynamics model $\phi_\theta$. The dynamics model predicts the added noise $\hat{\boldsymbol{\epsilon}} = \phi_\theta(\mathbf{z}_t, t)$ by alternating between equivariant coordinate updates and invariant feature messages across the 10 virtual latent nodes. The training objective minimizes the L2 loss between the true and predicted noise:
\begin{equation}
    \mathcal{L}_{diff} = \mathbb{E}_{t, \boldsymbol{\epsilon}} \left[ \frac{1}{30} \| \boldsymbol{\epsilon}_x - \hat{\boldsymbol{\epsilon}}_x \|^2 + \frac{1}{320} \| \boldsymbol{\epsilon}_h - \hat{\boldsymbol{\epsilon}}_h \|^2 \right]
\end{equation}
We employed an EGNN with 6 layers and 128 hidden dimensions, trained for 15 epochs using AdamW ($LR=2 \times 10^{-4}$) and Exponential Moving Average (EMA, decay 0.999) to enhance sampling stability.

\textbf{Atomic Count Regression.} Since the BFN decoder requires an explicit atom count $N_{atoms}$ to reconstruct the molecular graph from the latent state, we trained an auxiliary invariant GNN regressor. The regressor consists of 3 distance-based GNN layers that process the generated pair $(\mathbf{z}_x, \mathbf{z}_h)$. By computing rotational-invariant distances $d_{ij} = \| \mathbf{z}_{x,i} - \mathbf{z}_{x,j} \|$ as edge features, the model predicts the scalar $N_{atoms}$ via a global mean-max pooling readout. This ensures that \textit{de novo} generated latents are coupled with physically consistent atomic counts during the final decoding stage.

\textbf{Sampling and Assembly.} New linkers are sampled from the prior $\mathcal{N}(\mathbf{0}, \mathbf{I})$ via 1,000 steps of ancestral sampling. The resulting latent variables are then processed by the LinkerVAE decoder to produce all-atom 3D structures. Finally, these linkers are integrated into periodic frameworks using a flow-matching assembly pipeline \cite{mofflow} or standard algorithmic assembly tools \cite{tobacco}, completing the end-to-end MOF discovery workflow.

\section*{Data availability}

The datasets analysed in this study are derived from publicly available data provided by the MOFDiff~\cite{mofdiff} project (available at https://github.com/microsoft/MOFDiff and archived at https://zenodo.org/records/10806179). 
The processed data generated during the current study are available from the corresponding author upon reasonable request.

\section*{Code availability}

The code used in this study is available upon reasonable request from the corresponding author.

\section*{Acknowledgements}

The authors acknowledge financial support from the National Natural Science Foundation of China (Grant No. 52576233) and the Shenzhen Science and Technology Program (Grant No. KCXFZ20240903093459001).

\section*{Author contributions}

Dongxu Ji conceived the project and provided overall guidance. Chaoran Zhang performed the experiments, developed the code, and wrote the manuscript. Guangyao Li provided technical support and contributed to potential laboratory experiments.

\section*{Competing interests}

The authors declare no competing interests.

\section*{Additional information}

Supplementary information is available for this paper.

Correspondence and requests for materials should be addressed to jidongxu@cuhk.edu.cn.

\bibliography{sn-bibliography}

\clearpage
\onecolumn 

\begin{center}
    \Large \textbf{Supplementary Information for}\\
    \vspace{0.5cm}
    \large \textbf{LEGO-MOF: Equivariant Latent Manipulation for Editable, Generative, and Optimizable MOF Design}
\end{center}
\vspace{1cm}

\setcounter{section}{0}
\setcounter{figure}{0}
\setcounter{table}{0}
\setcounter{equation}{0}
\renewcommand{\thesection}{S\arabic{section}}
\renewcommand{\thefigure}{S\arabic{figure}}
\renewcommand{\thetable}{S\arabic{table}}
\renewcommand{\theequation}{S\arabic{equation}}

\begin{center}
    {\Large \textbf{Supplementary Information}}\\[1em]
\end{center}

\vspace{2em}

\section*{Supplementary Notes}
\subsection*{Supplementary Note 1: LinkerVAE Architecture and BFN Parameters}

The LinkerVAE is designed to bridge the gap between discrete molecular graphs and a continuous, differentiable latent manifold. The specific architectural details and hyperparameters are summarized below:

\textbf{1.1 SE(3)-Equivariant Encoder}
The encoder adopts an SE(3)-Transformer architecture to ensure that the learned latent representation is invariant to the global orientation of the input linker.
\begin{itemize}
    \item \textbf{Layers:} 9 Equivariant Attention layers.
    \item \textbf{Hidden Dimensions:} 128-dimensional node features.
    \item \textbf{Attention Heads:} 16 heads per layer.
    \item \textbf{Latent Bottleneck:} We utilize 10 virtual "global nodes" to aggregate information. These nodes are initialized with learnable embeddings and interact with the physical atom nodes through cross-attention.
    \item \textbf{Disentanglement:} The final state of these 10 nodes is split into $Z_h \in \mathbb{R}^{32}$ (invariant chemical features) and $Z_x \in \mathbb{R}^{10 \times 3}$ (equivariant spatial coordinates).
\end{itemize}

\textbf{1.2 Bayesian Flow Network (BFN) Decoder}
\begin{itemize}
    \item \textbf{Coordinate loss:} Modeled as a continuous flow with a precision parameter $\beta=1.0$.
    \item \textbf{Atom type loss:} Modeled as a discrete Bayesian flow over 17 classes (C, N, O, S, etc.).
    \item \textbf{Anchor Weighting:} During training, the reconstruction loss for virtual Helium atoms ($Z=2$) is multiplied by a factor of 1.5. This ensures the model prioritizes the accuracy of coordination trajectories over secondary structural details.
    \item \textbf{Sampling steps:} During inference, we utilize $K=100$ refinement steps to decode the molecular graph from the latent variables.
\end{itemize}

\subsection*{Supplementary Note 2: Equivariant Linker Latent Diffusion Model (LLDM) and Atomic Regressor}

The LLDM performs unconditional generation within the continuous $(Z_x, Z_h)$ manifold. The model architecture follows the Equivariant Variational Diffusion (EVD) framework.

\textbf{2.1 Noise Schedule and Forward Process}
The transition from the data distribution to the Gaussian prior is governed by a predefined polynomial noise schedule, as implemented in \texttt{polynomial\_schedule}:
\begin{itemize}
    \item \textbf{Timesteps ($T$):} 1000 steps.
    \item \textbf{Schedule type:} Polynomial with power $p=3.0$.
    \item \textbf{Precision ($s$):} $1 \times 10^{-4}$.
    \item \textbf{SNR Bounds:} The signal-to-noise ratio is bounded between $\log \text{SNR}_{max} \approx 5.0$ and $\log \text{SNR}_{min} \approx -10.0$ to ensure numerical stability.
\end{itemize}

\textbf{2.2 EGNN Dynamics Model}
The reverse denoising process is parameterized by an Equivariant Graph Neural Network (EGNN), which predicts the added noise $\hat{\epsilon}$ at each timestep $t$.
\begin{itemize}
    \item \textbf{Layers:} 6 layers of equivariant message passing.
    \item \textbf{Hidden NF:} 128 dimensions.
    \item \textbf{Time Conditioning:} The scalar timestep $t \in [0, 1]$ is embedded using a Sinusoidal Position Embedding ($dim=128$) and concatenated with the node features $Z_h$.
    \item \textbf{Equivariant Constraints:} The model strictly preserves the center of mass (CoM) of $Z_x$. Throughout the diffusion and sampling process, we enforce $\sum_{i=1}^{10} \mathbf{z}_{x,i} = 0$ via a masking operator to avoid translational drift.
    \item \textbf{Normalization:} We use a normalization factor of 100 for the coordinates within the EGNN to stabilize the radial basis interactions.
\end{itemize}

\textbf{2.3 Auxiliary Atom Count Regressor}
Since the BFN decoder requires a target atom count $N_{atoms}$, we use an auxiliary regressor to map generated latents back to discrete counts.
\begin{itemize}
    \item \textbf{Architecture:} 3-layer invariant Graph Neural Network.
    \item \textbf{Features:} Processes the Euclidean distance matrix derived from $Z_x$ and the invariant features from $Z_h$.
    \item \textbf{Readout:} Global mean-max pooling followed by an MLP with a ReLU activation at the final layer to ensure positive atom counts.
    \item \textbf{Accuracy:} Achieves a Mean Absolute Error (MAE) of 0.534 atoms on the validation set, ensuring that 92.7\% of generated linkers stay within $\pm 1$ atom of their physical capacity.
\end{itemize}

\subsection*{Supplementary Note 3: Property Surrogate}

\textbf{3.1 Periodic SchNet Surrogate}
The surrogate model for CO$_2$ uptake operates on a coarse-grained graph representation of the MOF.
\begin{itemize}
    \item \textbf{Cutoff Radius:} 35.0 \AA~ (calculated under Periodic Boundary Conditions).
    \item \textbf{RBF Kernels:} 50 kernels covering the range [0, 35] \AA~ for inter-linker distance encoding.
    \item \textbf{Message Passing:} 6 layers of continuous-filter convolution.
    \item \textbf{Conditioning:} Metal node types are embedded into a 32-dimensional space and integrated into the linker features using Feature-wise Linear Modulation (FiLM).
    \item \textbf{Global Pooling:} Attention-based pooling is used to aggregate the latent representations of all linkers in the unit cell.
\end{itemize}

\newpage
\section*{Supplementary Tables}

\begin{table}[htbp]
\centering
\caption{\textbf{Reconstruction performance of LinkerVAE on the test set.} Metrics are calculated between the ground truth and the sampled reconstruction. Anchor refers to the virtual connection atoms (He) required for MOF assembly.}
\label{tab:reconstruction}
\begin{tabular}{@{}lcc@{}}
\toprule
\textbf{Metric} & \textbf{Mean} & \textbf{Std. Dev.} \\
\midrule
Unaligned RMSD (\AA) & 0.5662 & 0.9115 \\
Kabsch Aligned RMSD (\AA) & 0.5662 & 0.9115 \\
Atom Type Accuracy (w/o anchors) & 94.84\% & 8.75\% \\
Atom Type Accuracy (all atoms) & 95.26\% & 8.10\% \\
Anchor Count Difference Error & \textbf{0.0000} & \textbf{0.0000} \\
Anchor Spatial Deviation (\AA) & \textbf{0.3072} & 0.5386 \\
\bottomrule
\end{tabular}
\end{table}

\vspace{2em}

\begin{table}[htbp]
\centering
\caption{\textbf{Evaluation of linker structures under atom count perturbations.} $\Delta$ represents the induced change in the total number of atoms during the decoding phase. All evaluations preserve a 100\% chemical validity.}
\label{tab:atomic_change}
\begin{tabular}{@{}lccccc@{}}
\toprule
\textbf{$\Delta$ (Atoms)} & \textbf{Samples} & \textbf{Validity} & \textbf{RMSD (\AA)} & \textbf{Anchor Num Diff} & \textbf{Anchor Dist (\AA)} \\
\midrule
-3 & 1000 & 100\% & -- & 0.031 $\pm$ 0.198 & 0.476 $\pm$ 0.692 \\
-2 & 1000 & 100\% & -- & 0.052 $\pm$ 0.289 & 0.437 $\pm$ 0.528 \\
-1 & 1000 & 100\% & -- & 0.014 $\pm$ 0.200 & 0.458 $\pm$ 0.678 \\
 0 (Recon)& 1000 & 100\% & 0.279 $\pm$ 0.360 & 0.016 $\pm$ 0.171 & 0.377 $\pm$ 0.596 \\
+1 & 1000 & 100\% & -- & 0.014 $\pm$ 0.210 & 0.442 $\pm$ 0.682 \\
+2 & 1000 & 100\% & -- & 0.031 $\pm$ 0.346 & 0.453 $\pm$ 0.678 \\
+3 & 1000 & 100\% & -- & 0.016 $\pm$ 0.215 & 0.439 $\pm$ 0.586 \\
\bottomrule
\end{tabular}
\end{table}

\vspace{2em}

\begin{table}[htbp]
\centering
\caption{\textbf{Quantitative metrics for structural style transfer.} $Z_h$ was mixed evenly (0.5/0.5) between parent pairs, while $Z_x$ was locked to either Parent 1 (Gen 1) or Parent 2 (Gen 2). Shape similarity is calculated via spatial Tanimoto overlap. Anchor distance tracks the spatial shift of connection nodes relative to the geometric parent.}
\label{tab:style_transfer}
\begin{tabular}{@{}lcccc@{}}
\toprule
\textbf{Generated Set} & \textbf{Geometric Parent} & \textbf{Validity} & \textbf{Shape Similarity} & \textbf{Anchor Dist (\AA)} \\
\midrule
Gen 1 (using $Z_{x,1}$) & Parent 1 & 100\% & 0.646 & 0.283 \\
Gen 2 (using $Z_{x,2}$) & Parent 2 & 100\% & 0.669 & 0.253 \\
\bottomrule
\end{tabular}
\end{table}

\vspace{2em}

\begin{table}[htbp]
\centering
\caption{\textbf{Quantitative physical property changes following zero-shot latent expansion (1.4$\times$).} Metrics represent the relative percentage shift from the original MOF to the optimized, expanded generated structure across all valid samples (88.15\% of total attempts). The absolute preservation of channel counts confirms the strict retention of topology.}
\label{tab:scaling_metrics}
\begin{tabular}{@{}lcccc@{}}
\toprule
\textbf{Property Metric} & \textbf{Mean Change (\%)} & \textbf{Std. Dev.} & \textbf{Min Change (\%)} & \textbf{Max Change (\%)} \\
\midrule
Unit Cell Volume & +38.66 & 11.92 & +2.89 & +75.01 \\
Framework Density & --11.45 & 10.82 & --48.01 & +25.45 \\
Accessible Vol. Fraction & +5.48 & 24.20 & --47.98 & +164.89 \\
Gravimetric AV (cm$^3$/g) & +23.32 & 43.27 & --49.45 & +289.32 \\
\midrule
Pore Channels Diff ($\Delta$) & \textbf{0.00} & \textbf{0.00} & \textbf{0.00} & \textbf{0.00} \\
\bottomrule
\end{tabular}
\end{table}
\vspace{2em}

\begin{table}[htbp]
\centering
\caption{\textbf{Summary of Test-Time Optimization (TTO) performance across three experiments.} Metrics are reported for MOFs that passed filtering criteria. Crucially, the mean and median relative boosts represent the \textbf{net overall shift}, calculated across \textit{all} valid samples in the given experiment (explicitly accounting for both performance improvements and degradations).}
\label{tab:tto_metrics}
\small 
\setlength{\tabcolsep}{3pt} 
\begin{tabular}{clcccc}
\toprule
\textbf{Exp.} & \begin{tabular}[c]{@{}l@{}}\textbf{Property}\\\textbf{Target}\end{tabular} & \begin{tabular}[c]{@{}c@{}}\textbf{Valid}\\\textbf{Samples}\end{tabular} & \begin{tabular}[c]{@{}c@{}}\textbf{Improved}\\\textbf{Ratio (\%)}\end{tabular} & \begin{tabular}[c]{@{}c@{}}\textbf{Net Mean}\\\textbf{Rel. Boost (\%)}\end{tabular} & \begin{tabular}[c]{@{}c@{}}\textbf{Net Median}\\\textbf{Rel. Boost (\%)}\end{tabular} \\
\midrule
1 & Pure CO$_2$ Uptake & 520 & 74.0 & +147.5 & +23.6 \\
2 & Binary CO$_2$/N$_2$ Uptake & 492 & 70.3 & +94.8 & +13.7 \\
3 & Top 0.33\% CO$_2$ Uptake & 908 & 44.9 & +1.1 & --2.6 \\
\bottomrule
\end{tabular}
\end{table}

\vspace{2em}

\begin{table}[htbp]
\centering
\caption{\textbf{Evaluation metrics for generated linkers.} Validity (V) denotes the fraction of generated samples that are successfully decoded into chemically valid molecules. Uniqueness (U) is defined as the proportion of non-redundant structures within the valid set, while Novelty (N) quantifies the percentage of unique molecules not present in the training database.}
\label{tab:vnu_metrics}
\begin{tabular}{@{}llc@{}}
\toprule
\textbf{Metric} & \textbf{Count} & \textbf{Percentage} \\
\midrule
Total Generated Samples & 1,000 & 100.00\% \\
Valid (V) & 975 & 97.50\% \\
Unique (U) & 777 & 79.69\% \textit{(of Valid)} \\
Novel (N) & 638 & 82.11\% \textit{(of Unique)} \\
\midrule
\textbf{Overall Unique \& Novel} & \textbf{638} & \textbf{63.80\% \textit{(Absolute)}} \\
\bottomrule
\end{tabular}
\end{table}

\vspace{2em}

\begin{table}[htbp]
\centering
\caption{\textbf{Overview of the MOF building block dataset and global atom distribution.} The dataset is decomposed into organic linkers and metal nodes. Atomic species are represented by their atomic number ($Z$). Note that $Z=2$ (Helium) denotes virtual anchor atoms, and $Z=1$ denotes Hydrogen.}
\label{tab:dataset_global_stats}
\footnotesize
\setlength{\tabcolsep}{3pt}
\begin{tabular}{@{}lcclcc@{}}
\toprule
\textbf{Dataset Metric} & \textbf{Count} & \textbf{Ratio} & \textbf{Atom Type ($Z$)} & \textbf{Total Count} & \textbf{Global Ratio (\%)} \\
\midrule
Total MOFs & 308,961 & -- & $Z=1$ (H) & 13,080,711 & 22.47 \\
Total Building Blocks (BBs) & 2,175,295 & 100\% & $Z=2$ (He) & 11,461,142 & 19.69 \\
Organic Linkers & 1,511,159 & 69.5\% & $Z=6$ (C) & 20,952,837 & 35.99 \\
Metal Nodes & 664,136 & 30.5\% & $Z=7$ (N) & 1,789,812 & 3.07 \\
\cmidrule{1-3}
\textbf{Metal Species Coverage} & \multicolumn{2}{l}{V, Cr, Ni, Cu, Zn, Ba} & $Z=8$ (O) & 8,211,113 & 14.11 \\
\textbf{Non-Metal Species} & \multicolumn{2}{l}{H, He, C, N, O, F, P, S, Cl, Br, I} & Others ($Z \ge 9$) & 2,751,210 & 4.67 \\
\bottomrule
\end{tabular}
\end{table}

\vspace{2em}

\begin{table}[htbp]
\centering
\caption{\textbf{Statistical analysis of organic linker building blocks.} (a) Distribution of atom counts per linker, indicating the variable-length nature of the inputs. (b) Relative abundance of chemical elements specifically within organic linkers.}
\label{tab:linker_stats}

\textbf{(a) Atom Count Distribution per Linker}\\[0.5em]
\begin{tabular}{@{}lccccccc@{}}
\toprule
\textbf{Metric} & \textbf{Mean} & \textbf{Median} & \textbf{Min} & \textbf{Max} & \textbf{P25} & \textbf{P75} & \textbf{P90} \\
\midrule
Atoms/Linker & 33.75 & 31.0 & 3 & 229 & 20.0 & 42.0 & 60.0 \\
\bottomrule
\end{tabular}

\vspace{1.5em}

\textbf{(b) Atomic Composition within Linkers}\\[0.5em]
\begin{tabular}{@{}lclc@{}}
\toprule
\textbf{Element ($Z$)} & \textbf{Ratio (\%)} & \textbf{Element ($Z$)} & \textbf{Ratio (\%)} \\
\midrule
$Z=6$ (Carbon) & 41.08 & $Z=2$ (Helium/Anchor) & 11.23 \\
$Z=1$ (Hydrogen) & 25.64 & $Z=7$ (Nitrogen) & 3.51 \\
$Z=8$ (Oxygen) & 15.87 & Halogens ($Z=9, 17, 35, 53$) & 1.97 \\
\bottomrule
\end{tabular}
\end{table}

\vspace{2em}

\begin{table}[htbp]
\centering
\caption{\textbf{Classification and frequency of metal node species.} Unique node species are identified by their core composition (excluding virtual He and H atoms). Frequencies reflect the occurrences across the 308,961 MOFs in the dataset.}
\label{tab:node_diversity}
\footnotesize
\setlength{\tabcolsep}{4pt}
\begin{tabular}{@{}cllcc@{}}
\toprule
\textbf{Rank} & \textbf{Node Core Composition} & \textbf{Common Structural Motif} & \textbf{Frequency} & \textbf{Percentage (\%)} \\
\midrule
1 & \{ Zn: 2 \} & Zn$_2$ dimer & 189,327 & 28.55 \\
2 & \{ Cu: 2 \} & Cu$_2$ paddlewheel & 177,294 & 26.73 \\
3 & \{ V: 1 \} & V Mononuclear & 91,484 & 13.80 \\
4 & \{ O: 1, Zn: 4 \} & Zn$_4$O cluster (basic zinc acetate) & 88,982 & 13.42 \\
5 & \{ Ni: 1 \} & Ni Mononuclear & 48,952 & 7.38 \\
6 & \{ Ba: 1 \} & Ba Mononuclear & 39,050 & 5.89 \\
7 & \{ O: 1, Cr: 3 \} & Cr$_3$O cluster & 29,000 & 4.37 \\
8 & Others (5 types) & Mononuclear or low-freq. clusters & 47 & $<$ 0.01 \\
\midrule
\multicolumn{2}{l}{\textbf{Node Core Stats (He/H removed)}} & \textbf{Mean}: 2.22 atoms/node & \textbf{P50}: 2.0 & \textbf{P90}: 5.0 \\
\bottomrule
\end{tabular}
\end{table}

\vspace{2em}

\begin{table}[htbp]
\centering
\caption{\textbf{Structural connectivity statistics.} Distribution of the number of neighboring metal nodes coordinated to a single organic linker across the framework library.}
\label{tab:connectivity_stats}
\begin{tabular}{@{}ccc@{}}
\toprule
\textbf{Neighboring Nodes ($k$)} & \textbf{Frequency (Linkers)} & \textbf{Percentage (\%)} \\
\midrule
2 & 1,321,408 & 87.44 \\
3 & 5,701 & 0.38 \\
4 & 142,170 & 9.41 \\
6 & 41,873 & 2.77 \\
Others & 7 & $<$ 0.01 \\
\midrule
\textbf{Mean Connectivity} & \multicolumn{2}{c}{2.30 nodes per linker} \\
\bottomrule
\end{tabular}
\end{table}

\clearpage 


\begin{figure}[htbp]
\centering
\begin{subfigure}{0.48\textwidth}
    \includegraphics[width=\textwidth]{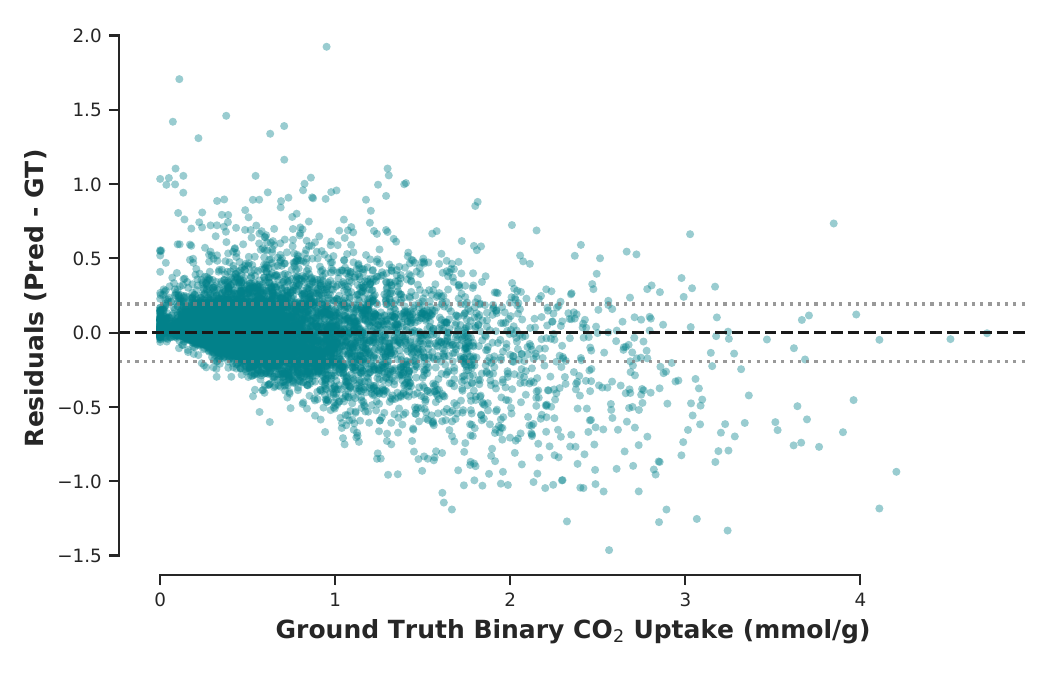}
    \caption{Unary $\text{CO}_2$ Uptake Residuals}
\end{subfigure}
\hfill
\begin{subfigure}{0.48\textwidth}
    \includegraphics[width=\textwidth]{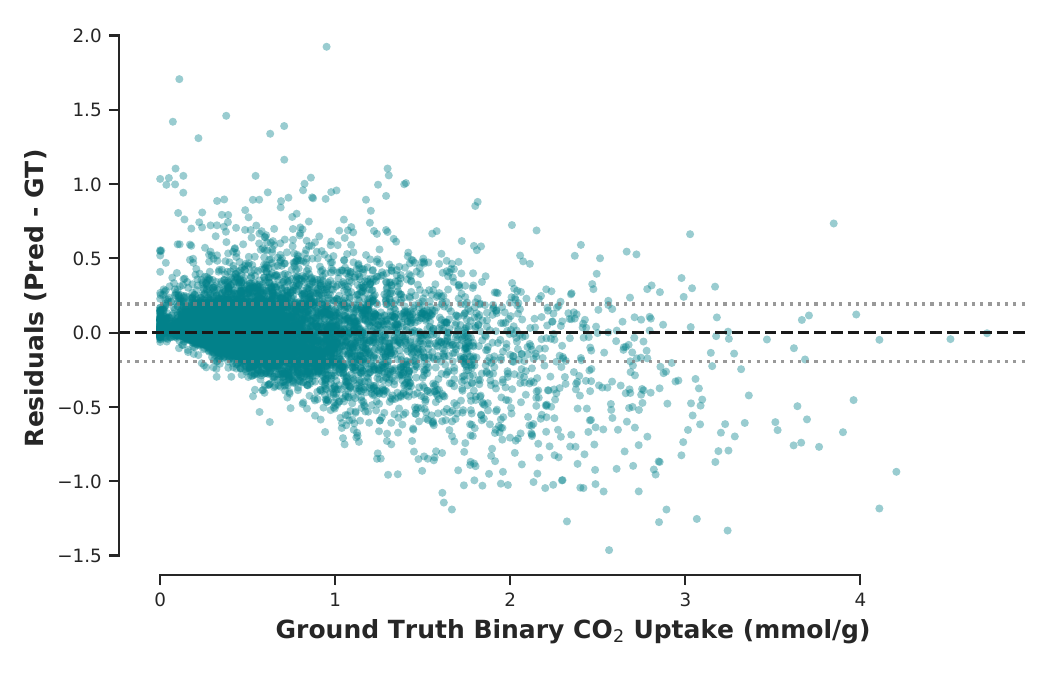}
    \caption{Binary $\text{CO}_2/\text{N}_2$ Uptake Residuals}
\end{subfigure}
\caption{\textbf{Residual analysis of the surrogate property predictor.} Scatter plots showing the prediction residuals (Predicted $-$ Ground Truth) as a function of the ground truth gas uptake for \textbf{a} pure (unary) $\text{CO}_2$ and \textbf{b} binary $\text{CO}_2/\text{N}_2$ mixture (0.15:0.85 bar). The dashed horizontal line represents zero error, while the dotted grey lines indicate the standard deviation bounds. The uniform distribution of residuals around zero across the entire range of uptake values indicates that the SchNet-based surrogate model is free from significant systematic bias and maintains high reliability even in high-performance regimes.}
\label{fig:supp_residuals}
\end{figure}

\vspace{3em}

\begin{figure}[htbp]
\centering
\includegraphics[width=0.75\textwidth]{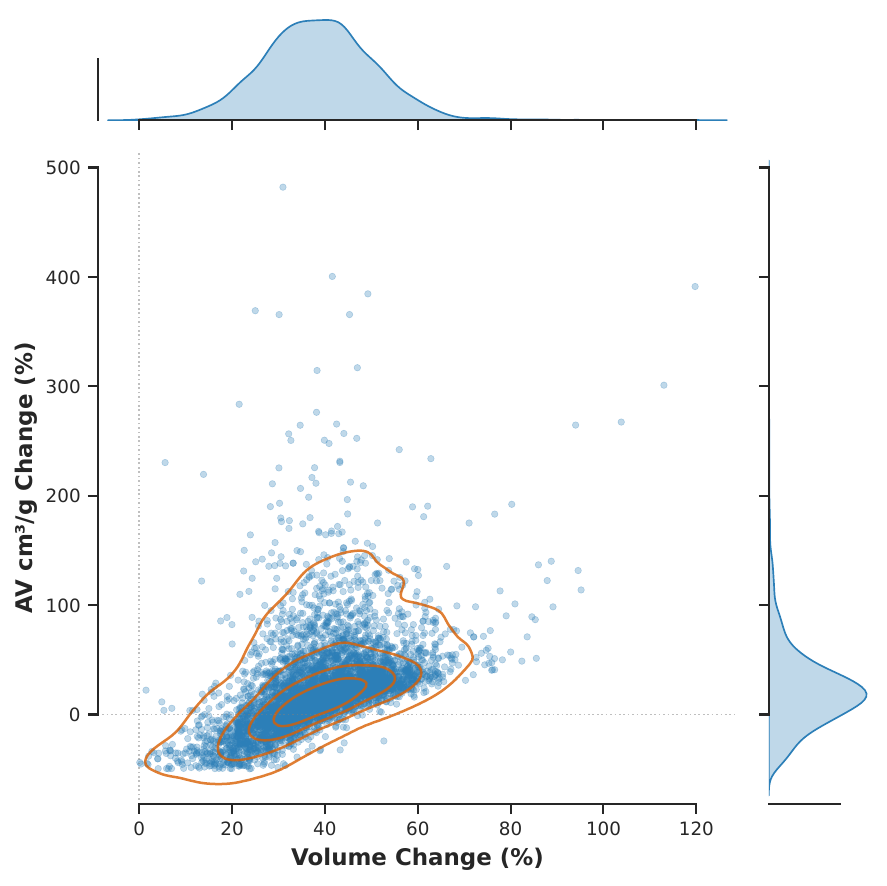}
\caption{\textbf{Detailed correlation between volumetric expansion and accessible volume (AV) enhancement.} Jointplot illustrating the relationship between the percentage change in unit cell volume and the resulting change in gravimetric AV ($\text{cm}^3/\text{g}$) following 1.4$\times$ latent geometric scaling. The scatter plot is overlaid with Kernel Density Estimation (KDE) contours, highlighting the high-density clustering of successfully expanded MOFs. The marginal histograms show the distribution for each metric, confirming a robust positive correlation where controlled macroscopic expansion leads to significant improvements in porous capacity across a diverse range of topologies.}
\label{fig:supp_jointplot}
\end{figure}

\end{document}